\documentclass{article}
\usepackage[preprint]{neurips_2019}
\usepackage[utf8]{inputenc} % allow utf-8 input
\usepackage[T1]{fontenc}    % use 8-bit T1 fonts
\usepackage{hyperref}       % hyperlinks
\usepackage{url}            % simple URL typesetting
\usepackage{booktabs}       % professional-quality tables
\usepackage{amsfonts}       % blackboard math symbols
\usepackage{nicefrac}       % compact symbols for 1/2, etc.
\usepackage{microtype}      % microtypography

\usepackage{url}
\usepackage{graphicx}
\usepackage{subfigure}
\usepackage{amsmath}
\usepackage{amssymb}
\usepackage{xcolor}

\usepackage{algorithm}
\usepackage[noend]{algorithmic}

\newtheorem{defi}{Definition}
\newtheorem{lemma}{Lemma}
\newtheorem{theorem}{Theorem}
\newtheorem{assume}{Assumption}
\newtheorem{coro}{Corollary}

\usepackage{mdframed}
\usepackage{lipsum}
\newmdtheoremenv{algo}{Algorithm}

\newlength\myindent
\title{Learning Privately over Distributed Features:\\An ADMM Sharing Approach}

\author{
 Yaochen Hu\\
 University of Alberta\\
 yaochen@ualberta.ca\\
 \And
 Peng Liu\\
 University of Kent\\
 P.Liu@kent.ac.uk
 \And
 Linglong Kong\\
 University of Alberta\\
 lkong@ualberta.ca\\
 \And
 Di Niu\\
 University of Alberta\\
 dniu@ualberta.ca
}

\begin{document}

\maketitle

\begin{abstract}
Distributed machine learning has been widely studied in order to handle exploding amount of data.
In this paper, we study an important yet less visited distributed learning problem where features are inherently distributed or vertically partitioned among multiple parties, and sharing of raw data or model parameters among parties is prohibited due to privacy concerns.
We propose an ADMM sharing framework to approach risk minimization over distributed features, where each party only needs to share a single value for each sample in the training process, thus minimizing the data leakage risk.
We establish convergence and iteration complexity results for the proposed parallel ADMM algorithm under non-convex loss. We further introduce a novel differentially private ADMM sharing algorithm and bound the privacy guarantee with carefully designed noise perturbation. The experiments based on a prototype system shows that the proposed ADMM algorithms converge efficiently in a robust fashion, demonstrating advantage over gradient based methods especially for data set with high dimensional feature spaces.
\end{abstract}

\section{Introduction}

%Distributed and collaborative machine learning has been widely studied in order to handle exploding amount of data.

The effectiveness of a machine learning model does not only depend on the quantity of samples, but also the quality of data, especially the availability of high-quality features.
Recently, a wide range of distributed and collaborative machine learning schemes, including gradient-based methods \cite{li2014scaling, li2014communication, hsieh2017gaia, ho2013more} and ADMM-based methods \cite{zhang2018improving, shi2014linear, zhang2016dual, huang2018dp},
have been proposed to enable learning from distributed samples, since collecting data for centralized learning will incur compliance overhead, privacy concerns, or even judicial issues. Most existing schemes, however, are under the umbrella of \emph{data parallel} schemes, where multiple parties possess different training samples, each sample with the same set of features. 
%are under the umbrella of \emph{data parallel} schemes.
For example, different users hold different images to jointly train a classifier.

An equally important scenario is to collaboratively learn from distributed features, where multiple parties may possess different features about a same sample, yet do not wish to share these features with each other. Examples include a user's behavioural data logged by multiple apps, a patient's record stored at different hospitals and clinics, a user's investment behavior logged by multiple financial institutions and government agencies and so forth. The question is---how can we train a joint model to make predictions about a sample leveraging the potentially rich and vast features possessed by other parties, without requiring different parties to share their data to each other?

The motivation of gleaning insights from vertically partitioned data dates back to association rule mining \cite{vaidya2002privacy,vaidya2003privacy}. A few very recent studies \cite{kenthapadi2013privacy,ying2018supervised, hu2019fdml, heinze2017preserving,dai2018privacy,stolpe2016sustainable} have reinvestigated vertically partitioned features under the setting of distributed machine learning, which is motivated by the ever-increasing data dimensionality as well as the opportunity and challenge of cooperation between multiple parties that may hold different aspects of information about the same samples. %These studies also differ from model parallelism \cite{zhou2016convergence}

In this paper, we propose an ADMM algorithm to solve the empirical risk minimization (ERM) problem, a general optimization formulation of many machine learning models visited by a number of recent studies on distributed machine learning  \cite{ying2018supervised, chaudhuri2011differentially}. We propose an ADMM-sharing-based distributed algorithm to solve ERM, in which each participant does not need to share any raw features or local model parameters to other parties. Instead, each party only transmits a single value for each sample to other parties, thus largely preventing the local features from being disclosed. We establish theoretical convergence guarantees and iteration complexity results under the non-convex loss in a fully parallel setting, whereas previously, the convergence of ADMM sharing algorithm for non-convex losses is only known for the case of sequential (Gauss-Seidel) execution \cite{hong2016convergence}.

To further provide privacy guarantees, we present a privacy-preserving version of the ADMM sharing algorithm, in which the transmitted value from each party is perturbed by a carefully designed Gaussian noise to achieve
the notion of $\epsilon,\delta$-differential privacy \cite{dwork2008differential, dwork2014algorithmic}. For distributed features, the perturbed algorithm ensures that the probability distribution of the values shared is relatively insensitive to any change to a single feature in a party's local dataset.
  %We theoretically show that the privacy cost of executing $T$ epochs of the algorithm is ... \red{say something interesting here}.

Experimental results on two realistic datasets suggest that our proposed ADMM sharing algorithm can converge efficiently. Compared to the gradient-based method, our method can scale as the number of features increases and yields robust convergence. The algorithm can also converge with moderate amounts of Gaussian perturbation added, therefore enabling the utilization of features from other parties to improve the local machine learning task.
%It has also been reported that the volume and quality of data determines the upper bound of machine learning model performance.
%As a result, a number of distributed machine learning techniques have been proposed to collaboratively train a model by letting each party perform local model updates and exchange locally computed gradients  \cite{shokri2015privacy} or model parameters \cite{mcmahan2016communication} with the central server to iteratively improve model accuracy. Most of the existing schemes, however, fall into the range of \emph{data parallel} computation, where the training samples are located on different parties. For example, different users hold different images to jointly train a classifier. Different organizations may contribute their individual corpora to learn a joint language model.
\subsection{Related Work}
{\bf Machine Learning Algorithms and Privacy.} \cite{chaudhuri2009privacy}
is one of the first studies combing machine learning and differential privacy (DP), focusing on logistic regression. \cite{shokri2015privacy} applies a variant of SGD to collaborative deep learning in a data-parallel fashion and introduces its variant with DP. \cite{abadi2016deep} provides a stronger differential privacy guarantee for training deep neural networks using a momentum accountant method. \cite{pathak2010multiparty, rajkumar2012differentially} apply DP to collaborative machine learning, with an inherent tradeoff between the privacy cost and utility achieved by the trained model. Recently, DP has been applied to ADMM algorithms to solve multi-party machine learning problems \cite{zhang2018improving, zhang2016dual, zhang2019admm, zhang2017dynamic}. 

However, all the work above is targeting the data-parallel scenario, where samples are distributed among nodes. The uniqueness of our work is to enable privacy-preserving machine learning among nodes with vertically partitioned features, or in other words, the feature-parallel setting, which is equally important and is yet to be explored. 

%Aside from the distributed optimization approach mentioned above, 
Another approach to privacy-preserving machine learning is through
encryption \cite{gilad2016cryptonets,takabi2016privacy, kikuchi2018privacy} or secret sharing \cite{mohassel2017secureml, wan2007privacy, bonte2018privacy}, so that models are trained on encrypted data. However, encryption cannot be generalized to all algorithms or operations, and incurs additional computational cost.

{\bf Learning over Distributed Features.}
\cite{gratton2018distributed} applies ADMM to solve ridge regression.
\cite{ying2018supervised} proposes a stochastic learning method via variance reduction. \cite{zhou2016convergence} proposes a proximal gradient method and mainly focuses on speeding up training in a model-parallel scenario. These studies do not consider the privacy issue. \cite{hu2019fdml} proposes a composite model structure that can jointly learn from distributed features via a SGD-based algorithm and its DP-enabled version, yet without offering theoretical privacy guarantees. Our work establishes the first $(\epsilon,\delta)$-differential privacy guarantee result for learning over distributed features. Experimental results further suggest that our ADMM sharing method converges in fewer epochs than gradient methods in the case of high dimensional features. This is critical to preserving privacy in machine learning since the privacy loss increases as the number of epochs increases \cite{dwork2014algorithmic}.  %is proportional to $O\sqrt{T}$, T being the number of epochs.

{\bf Querying Vertically Partitioned Data Privately.} \cite{vaidya2002privacy, evfimievski2004privacy, dwork2004privacy} are among the early studies that investigate the privacy issue of querying vertically partitioned data. \cite{kenthapadi2012privacy} adopts a random-kernel-based method to mine vertically partitioned data privately. These studies provide privacy guarantees for simpler static queries, while we focus on machine learning jobs, where the risk comes from the shared values in the optimization algorithm. Our design simultaneously achieves minimum message passing, fast convergence, and a theoretically bounded privacy cost under the DP framework.

\section{Empirical Risk Minimization over Distributed Features}
\label{sec:problem}

Consider $N$ samples, each with $d$ features distributed on $M$ parties, which do not wish to share data with each other.
The entire dataset $\mathcal D\in \mathbb{R}^N\times\mathbb{R}^{d}$ can be viewed as $M$ vertical partitions $\mathcal D_1,\ldots,\mathcal D_M$, where
$\mathcal{D}_m\in\mathbb{R}^N\times\mathbb{R}^{d_m}$ denotes the data possessed by the $m$th party and $d_m$ is the dimension of features on party $m$. Clearly, $d=\sum_{m=1}^{M}d_m$. Let $\mathcal{D}^i$ denote the $i$th row of $\mathcal{D}$, and
$\mathcal{D}_m^i$ be the $i$th row of $\mathcal{D}_m$ ($k=1,\cdots,N$). Then, we have
\begin{eqnarray*}
	\mathcal{D} =
\left[
  \begin{array}{cccc}
    \mathcal{D}_1^1 & \mathcal{D}_2^1 & \cdots & \mathcal{D}_M^1 \\
    \mathcal{D}_1^2 & \mathcal{D}_2^2 &\cdots & \mathcal{D}_M^2 \\
    \vdots & \vdots & \ddots & \vdots \\
    \mathcal{D}_1^N & \mathcal{D}_2^N & \cdots & \mathcal{D}_M^N
  \end{array}
\right],
\end{eqnarray*}
where $\mathcal{D}_m^i\in\mathcal{A}_m\subset\mathbb{R}^{d_m}$, ($i=1,\cdots,N, m=1,\cdots,M$).
Let $Y_i\in\{-1, 1\}^N$ be the label of sample $i$. %Then, $(\mathcal{D}_1, \mathcal{D}_2, \ldots, \mathcal{D}_M, Y)$ is the total training data.

Let $x=(x_1^\top,\cdots,x_m^\top,\cdots,x_M^\top)^\top$ represent the model parameters, where $x_m\in\mathbb{R}^{d_m}$ are the local parameters associated with the $m$th party. The objective is to find a model $f(\mathcal{D}^i; x)$ with parameters $x$ %to predict the labels such that some predefined loss $l(f, Y)$ between the predicted labels and true labels is minimized, i.e.,
to minimize the regularized empirical risk, i.e.,
\[
\underset{x \in X}{\text{minimize}} \quad\frac{1}{N}\sum_{i=1}^{N} l_i(f(\mathcal{D}^i; x), Y_i) + \lambda R(x),
\]
where $X\subset\mathbb{R}^d$ is a closed convex set and the regularizer $R(\cdot)$ prevents overfitting.
% We consider \emph{generalized additive models}
%We consider models that linearly depends on the features
% \begin{align}
% f(D_1^i, D_2^i, \ldots, D_m^i, x) = \sigma(\sum_j h_j(D_j^i, x_j)),
% \end{align}

Similar to recent literature on distributed machine learning \cite{ying2018supervised, zhou2016convergence}, ADMM \cite{zhang2016dual, zhang2018improving}, and privacy-preserving machine learning \cite{chaudhuri2011differentially, hamm2016learning}, we assume the loss has a form
\[
\sum_{i=1}^{N}l_i(f(\mathcal{D}^i; x), Y_i) = \sum_{i=1}^{N}l_i(\mathcal{D}^i x, Y_i)
=l\left(\sum_{m=1}^{M} \mathcal{D}_m^i x_m\right),
\]
% where $h_j(\cdot, \cdot)$ is some local aggregation function for the features on party $j$
where we have abused the notation of $l$ and in the second equality absorbed the label $Y_i$ into the loss $l$, which is possibly a non-convex function.
This framework incorporates a wide range of commonly used models including support vector machines, Lasso, logistic regression, boosting, etc.

%This class of functions include a wide range of realistic models such as lasso, logistic regression, SVM, etc.Taking logistic regress as example, we have
%\begin{align}
%f(\mathcal{D}_1^i, \mathcal{D}_2^i, \ldots, \mathcal{D}_M^i, x) = 1/(1+\text{exp}(-\sum_{m=1}^{M} \mathcal{D}_m^ix_m)), i=1,2,\cdots,N
%\end{align}
%and
%\begin{align}
%l_i(f_i, Y_i) = & \text{log}(1+\text{exp}(-Y_i\sum_{m=1}^{M} \mathcal{D}_m^ix_m)), i=1,2,\cdots,N.
%\end{align}

Therefore, the risk minimization over distributed features, or vertically partitioned datasets $\mathcal D_1,\ldots,\mathcal D_M$, can be written in the following compact form:
\begin{align}
\underset{x}{\text{minimize}}&\quad l\left(\sum_{m=1}^{M} \mathcal{D}_mx_m\right) + \lambda\sum_{m=1}^{M} R_m(x_m), \label{eq:analysis_problem}\\
\text{subject to}&\quad x_m\in X_m, m=1,\ldots,M,
\end{align}
where  $X_m\subset\mathbb{R}^{d_m}$ is a closed convex set for all $m$.

We have further assumed the regularizer is separable such that
$R(x) = \sum_{m=1}^{M} R_m(x_m).$ This assumption is consistent with our algorithm design philosophy---under vertically partitioned data, we require each party focus on training and regularizing its local model $x_m$, without sharing any local model parameters or raw features to other parties at all. 
\section{The ADMM Sharing Algorithm}
We present an ADMM sharing algorithm \cite{boyd2011distributed, hong2016convergence} to solve Problem~\eqref{eq:analysis_problem} and establish a convergence guarantee for the algorithm. Our algorithm requires each party only share a single value to other parties in each iteration, thus requiring the minimum message passing. 
In particular, Problem~\eqref{eq:analysis_problem} is equivalent to
\begin{align}
\underset{x}{\text{minimize}} &\quad l\left(z\right) + \lambda\sum_{m=1}^{M} R_m(x_m),\\
\text{s.t.} &\quad \sum_{m=1}^{M} \mathcal{D}_m x_m - z = 0,\quad x_m\in X_M, m=1,\ldots,M,
\end{align}
where $z$ is an auxiliary variable. 
% Define $\{x\}=\{x_1,\cdots,x_M\}$.
The corresponding augmented Lagrangian is given by
\begin{align}
\mathcal{L}(\{x\}, z; y) = l(z) + \lambda\sum_{m=1}^{M} R_m(x_m) + \langle y, \sum_{m=1}^{M}\mathcal{D}_m x_m - z\rangle + \frac{\rho}{2}\|\sum_{m=1}^{M} \mathcal{D}_m x_m - z\|^2, \label{eq:lagragian}
\end{align}
where $y$ is the dual variable and $\rho$ is the penalty factor.
In the $t^{th}$ iteration of the algorithm, variables are updated according to
% {\bf the sequential updating algorithm}
% \begin{align}
% &x_j^{t+1}:=\underset{x_j\in X_j}{\text{argmin}}\quad\lambda R_j(x_j) + \langle y^t, D_jx_j\rangle + \frac{\rho}{2}\big\|\sum_{k<j}D_kx_k^{t+1} + \sum_{k>j}D_kx_k^t + D_jx_j - z^t\big\|^2\label{eq:seq_algo_x}\\
% &z^{t+1}:=\underset{z}{\text{argmin}}\quad l(z) - \langle y^t, z \rangle + \frac{\rho}{2} \big\|\sum_jD_jx_j^{t+1} - z\big\|^2\label{eq:seq_algo_z}\\
% &y^{t+1}:=y^t + \rho\big(\sum_jD_jx_j^{t+1} - z^{t+1}\big),\label{eq:seq_algo_y}
% \end{align}
% and  {\bf the parallel updating algorithm}

\begin{align}
&x_m^{t+1}:=\underset{x_m\in X_m}{\text{argmin}}\quad\lambda R_m(x_m) + \langle y^t, \mathcal{D}_mx_m\rangle + \frac{\rho}{2}\big\|\sum_{\substack{k=1,~k\neq m}}^{M}\mathcal{D}_kx_k^{t} + \mathcal{D}_mx_m - z^t\big\|^2, \nonumber\\&\hspace*{36pt} m=1,\cdots,M\label{eq:pal_algo_x}\\
&z^{t+1}:=\underset{z}{\text{argmin}}\quad l(z)  - \langle y^t, z \rangle + \frac{\rho}{2} \big\|\sum_{m=1}^{M}\mathcal{D}_mx_m^{t+1} - z\big\|^2\label{eq:pal_algo_z}\\
&y^{t+1}:=y^t + \rho\big(\sum_{m=1}^{M}\mathcal{D}_mx_m^{t+1} - z^{t+1}\big).\label{eq:pal_algo_y}
\end{align}

Formally, in a distributed and fully parallel manner, the algorithm is described in Algorithm~\ref{alg:ADMM_sharing}. Note that each party $m$ needs the value $\sum_{k\neq m}\mathcal{D}_kx_k^{t} - z^{t}$ to complete the update, and Line~\ref{alg:line_1}, \ref{alg:line_2} and \ref{alg:line_8} in Algorithm~\ref{alg:ADMM_sharing} present a trick to reduce communication overhead.

\begin{algorithm}[t]
\caption{The ADMM Sharing Algorithm}
\begin{algorithmic}[1]
    \STATE -----\emph{Each party $m$ performs in parallel:}
    % \bindent
    \FOR {$t$ in $1, \ldots, T$}
        \STATE Pull  $\sum_k\mathcal{D}_kx_k^{t} - z^{t}$  and $y^{t}$ from central node \label{alg:line_1}
        \STATE Obtain $\sum_{k\neq m}\mathcal{D}_kx_k^{t} - z^{t}$ by subtracting the locally cached $\mathcal{D}_mx_m^{t}$ from  the pulled value $\sum_k\mathcal{D}_kx_k^{t} - z^{t}$ \label{alg:line_2}
        \STATE Compute $x_m^{t+1}$ according to \eqref{eq:pal_algo_x} \label{alg:line_3}
        \STATE Push $\mathcal{D}_mx_m^{t+1}$ to the central node \label{alg:line_4}
    \ENDFOR
    % \eindent
    \STATE -----\emph{Central node:}
    % \bindent
    \FOR{$t$ in $1, \ldots, T$}
        \STATE Collect $\mathcal{D}_mx_m^{t+1}$ for all $m=1,\ldots,M$\label{alg:line_5}
        \STATE Compute $z^{t+1}$ according to \eqref{eq:pal_algo_z}\label{alg:line_6}
        \STATE Compute $y^{t+1}$ according to \eqref{eq:pal_algo_y}\label{alg:line_7}
        \STATE Distribute $\sum_k\mathcal{D}_kx_k^{t+1} - z^{t+1}$  and $y^{t+1}$ to all the parties. \label{alg:line_8}
    \ENDFOR
    % \eindent
\end{algorithmic}
\label{alg:ADMM_sharing}
\end{algorithm}

% \subsection{Privacy Concern and Differential Privacy}

% Random noise added to the query. Variance is correlated to the DP guarantee.
% {\bf Laplace Mechanism}
% \begin{align}
% \eta\sim\text{Lap}(\frac{\Delta f}{\epsilon})
% \end{align}

% ADMM sharing with differential privacy.
% \begin{align}
% &x_m^{t+1}:=\underset{x_m\in X_m}{\text{argmin}}\quad\lambda R_m(x_m) + \langle y^t, \mathcal{D}_mx_m\rangle + \frac{\rho}{2}\big\|\sum_{\substack{k=1\\k\neq m}}^M\mathcal{D}_kx_k + \mathcal{D}_mx_m - z^t\big\|^2\text{~for~all}~ m=1,\cdots,M\label{eq:pri_algo_x}\\
% & \text{Generate~} \xi_m^{t+1}\sim\mathcal{N}(0, \sigma^2_{m, t+1})\text{~for~all}~ m=1,\cdots,M\label{eq:pri_algo_xi}\\
% %%%&Q_m^{t+1} = \mathcal{D}_mx_m^{t+1} + \xi_m^{t+1} \text{~for~all}~ m=1,\cdots,M\label{eq:pri_algo_M} \\
% &\tilde{x}_m^{t+1}\leftarrow x_m^{t+1}+\xi_m^{t+1}\text{~for~all}~ m=1,\cdots,M\label{eq:pri_algo_M} \\
% &z^{t+1}:=\underset{z}{\text{argmin}}\quad l(z)  - \langle y^t, z \rangle + \frac{\rho}{2} \big\|\sum_{m=1}^{M}\mathcal{D}_m\tilde{x}_m^{t+1} - z\big\|^2\label{eq:pri_algo_z}\\
% &y^{t+1}:=y^t + \rho\big(\sum_{m=1}^{M}\mathcal{D}_m\tilde{x}_m^{t+1} - z^{t+1}\big).\label{eq:pri_algo_y}
% \end{align}

\subsection{Convergence Analysis}
We follow Hong et al. \cite{hong2016convergence} to establish the convergence guarantee of the proposed algorithm under mild assumptions. Note that \cite{hong2016convergence} provides convergence analysis for the Gauss-Seidel version of the ADMM sharing, where $x_1,\ldots,x_M$ are updated sequentially, which is not naturally suitable to parallel implementation.
In~\eqref{eq:pal_algo_x} of our algorithm, $x_m$'s can be updated by different parties in parallel in each iteration.
We establish convergence as well as iteration complexity results for this parallel scenario, which is more realistic in distributed learning. We need the following set of common assumptions. 
\begin{assume}\label{theo:assumptions_pri}
% \newline
\begin{enumerate}
    \item There exists a positive constant $L>0$ such that
        \[
            \|\nabla l(x)-\nabla l(z)\| \le L\|x-z\|\quad \forall x, z.
        \]
        Moreover, for all $m\in\{1,2,\cdots,M\}$, $X_m$'s are closed convex sets; each $\mathcal{D}_m$ is of full column rank so that the minimum eigenvalue $\sigma_{\text{min}}(\mathcal{D}_m^\top \mathcal{D}_m)$ of matrix $\mathcal{D}_m^\top \mathcal{D}_m$ is positive.\label{item:assum_1_pri}
    \item The penalty parameter $\rho$ is chosen large enough such that
    \begin{enumerate}
        \item each $x_m$ subproblem~\eqref{eq:pal_algo_x} as well as the $z$ subproblem~\eqref{eq:pal_algo_z} is strongly convex, with modulus $\{\gamma_m(\rho)\}_{m=1}^M$ and $\gamma(\rho)$, respectively. \label{item:asusum_2_1_pri}
        \item $\gamma_m(\rho)\ge 2\sigma_{\text{max}}(\mathcal{D}_m^\top \mathcal{D}_m), \forall m$, where $\sigma_{\text{max}}(\mathcal{D}_m^\top \mathcal{D}_m)$ is the maximum eigenvalue for matrix $\mathcal{D}_m^\top \mathcal{D}_m$.
        \item$\rho\gamma(\rho)>2L^2$ and $\rho\ge L$.
    \end{enumerate}
    \label{item:assum_2_pri}
    \item The objective function $l\left(\sum_{m=1}^{M} \mathcal{D}_mx_m\right) + \lambda\sum_{m=1}^{M} R_m(x_m)$ in Problem~\ref{eq:analysis_problem} is lower bounded over $\Pi_{m=1}^MX_m$ and we denote the lower bound as $\underline{f}$.\label{item:assum_3_pri}
    \item $R_m$ is either smooth nonconvex or convex (possibly nonsmooth). For the former case, there exists $L_m>0$ such that $\|\nabla R_m(x_m) - \nabla R_m(z_m)\|\le L_m\|x_m-z_m\|$ for all $x_m, z_m\in X_m$.\label{item:assum_4_pri}
\end{enumerate}
\end{assume}
Specifically, \ref{item:assum_1_pri}, \ref{item:assum_3_pri} and \ref{item:assum_4_pri} in Assumptions~\ref{theo:assumptions_pri} are common settings in the literature. Assumptions~\ref{theo:assumptions_pri}.\ref{item:assum_2_pri} is achievable if the $\rho$ is chosen large enough.

Denote $\mathcal{M}\subset\{1,2,\ldots, M\}$ as the index set, such that when $ m\in\mathcal{M}$, $R_m$ is convex, otherwise, $R_m$ is nonconvex but smooth. Our convergence results show that under mild assumptions, the iteratively updated variables eventually converge to the set of primal-dual stationary solutions. 
\begin{theorem}\label{theo:convergence}
Suppose Assumption~\ref{theo:assumptions_pri} holds true, we have the following results:
\begin{enumerate}
    \item $\lim_{t\rightarrow\infty}\|\sum_{m=1}^{M} \mathcal{D}_mx_m^{t+1} - z^{t+1}\|$=0.\label{item:primal_cond_limit}
    \item Any limit point $\{\{x^*\}, z^*; y^*\}$ of the sequence $\{\{x^{t+1}\}, z^{t+1}; y^{t+1}\}$ is a stationary solution of problem~\eqref{eq:analysis_problem} in the sense that
    \begin{align}
        & x_m^* \in \underset{x_m\in X_m}{\text{argmin}}\quad \lambda R_m(x_m) + \langle y^*, \mathcal{D}_mx_m\rangle, m\in\mathcal{M},\label{eq:cond_x_opt_conv}\\
        & \langle x_m - x_m^*, \lambda\nabla l(x_m^*) - \mathcal{D}_m^T y^* \rangle\le 0\quad\forall x_m\in X_m, m\not\in\mathcal{M}, \label{eq:cond_x_opt_nonconv}\\
        & \nabla l(z^*) - y^* = 0,\label{eq:cond_dual}\\
        & \sum_{m=1}^{M}\mathcal{D}_mx_m^* = z^*.\label{eq:cond_primal}
    \end{align}
    \item If $\mathcal{D}_m$ is a compact set for all $m$, then $\{\{x_m^t\}, z^t; y^t\}$ converges to the set of stationary solutions of problem~\eqref{eq:analysis_problem}, i.e.,
    \begin{align}
        \underset{t\rightarrow\infty}{\lim}\quad\text{dist}\big((\{x^t\}, z^t; y^t);Z^*\big) = 0,\nonumber
    \end{align}
    where $Z^*$ is the set of primal-dual stationary solutions for problem~\eqref{eq:analysis_problem}.
\end{enumerate}
\end{theorem}

\subsection{Iteration Complexity Analysis}
We evaluate the iteration complexity over a \emph{Lyapunov function}. More specifically, we define $V^t$ as
\begin{align}
    V^t:=\sum_{m=1}^{M} \|\tilde{\nabla}_{x_m} \mathcal{L}(\{x_m^t\}, z^t; y^t)\|^2 + \|\nabla_z \mathcal{L}(\{x_m^t\}, z^t; y^t)\|^2 + \|\sum_{m=1}^{M} \mathcal{D}_mx_m^t - z^t\|^2,\label{eq:Lyapunov}
\end{align}
where
\begin{align}
    & \tilde{\nabla}_{x_m} \mathcal{L}(\{x_m^t\}, z^t; y^t) = \nabla_{x_m} \mathcal{L}(\{x_m^t\}, z^t; y^t)\quad\hfill\text{when}~ m\not\in\mathcal{M},\nonumber\\
    & \tilde{\nabla}_{x_m} \mathcal{L}(\{x_m^t\}, z^t; y^t) = x_m^t - \text{prox}_{\lambda R_m} \big[x_m^t-\nabla_{x_m}\big(\mathcal{L}(\{x_m^t\}, z^t; y^t) - \lambda\sum_{m=1}^{M} R_m(x_m^t)\big)\big]\nonumber\\ 
    &\quad\hfill \text{when} ~m\in\mathcal{M},\nonumber
\end{align}
with $\text{prox}_h[z] := \text{argmin}_x h(x)+\frac{1}{2}\|x-z\|^2$. It is easy to verify that when $V^t\rightarrow 0$, a stationary solution is achieved due to the properties. The result for the iteration complexity is stated in the following theorem, which provides a quantification of how fast our algorithm converges. Theorem~\ref{theo:iter_complexity} shows that the algorithm converges in the sense that the \emph{Lyapunov function} $V^t$ will be less than any $\epsilon>0$ within $O(1/\epsilon)$ iterations. 
\begin{theorem}\label{theo:iter_complexity}
    Suppose Assumption~\ref{theo:assumptions_pri} holds. Let $T(\epsilon)$ denote the iteration index in which:
    \begin{align}
        T(\epsilon):=\text{min}\{t|V^t\le\epsilon, t\ge0\},\nonumber
    \end{align}
    for any $\epsilon>0$. Then there exists a constant $C>0$, such that
    \begin{align}
        T(\epsilon)\epsilon\le C(\mathcal{L}(\{x^1\}, z^1; y^1 - \underline{f}),
    \end{align}
    where $\underline{f}$ is the lower bound defined in Assumption~\ref{theo:assumptions_pri}.%\ref{item:assum_3_pri}.
\end{theorem} 
\section{Differentially Private ADMM Sharing}

Differential privacy \cite{dwork2014algorithmic, zhou2010security} is a notion that ensures a strong guarantee for data privacy. The intuition is to keep the query results from a dataset relatively close if one of the entries in the dataset changes, by adding some well designed random noise into the query, so that little information on the raw data can be inferred from the query. Formally, the definition of differential privacy is given in Definition~\ref{defi:DP}.
% The novelty of our proof is that we provide a differential privacy guarantee of
% $\mathcal{D}_m x^{t+1}_m$ --- the value shared to other parties, instead of $x^{t+1}_m$ itself.
\begin{defi}\label{defi:DP}
A randomized algorithm $\mathcal{M}$ is $(\varepsilon, \delta)-$differentially private if for all $S\subset\text{range}(\mathcal{M})$, and for all $x$ and $y$, such that $|x-y|_1\le 1$:
\begin{align}
\text{Pr}(\mathcal{M}(x))\le \exp(\varepsilon)\text{Pr}(\mathcal{M}(y))+\delta.
\end{align}
\end{defi}
%%%\begin{defi}
%%%$l_1-$sensitivity of function $f$ is
%%%\begin{align}
%%%\Delta f = \underset{\|x-y\|_1\le1}{\max}\|f(x)-f(y)\|_1.
%%%\end{align}
%%%\end{defi}
In our ADMM algorithm, the shared messages $\{\mathcal{D}_mx_m^{t+1}\}_{t=0,1,\cdots,T-1}$ may reveal sensitive information from the data entry in $D_m$ of Party $m$. We perturb the shared value $\mathcal{D}_mx^{t+1}_m$ in Algorithm~\ref{alg:ADMM_sharing} with a carefully designed random noise to provide differential privacy. The resulted perturbed ADMM sharing algorithm is the following updates:
\begin{align}
  &x_{m}^{t+1}:=\underset{x_m\in X_m}{\text{argmin}}\quad\lambda R_m(x_m) + \langle y^t, \mathcal{D}_mx_m\rangle + \frac{\rho}{2}\big\|\sum_{\substack{k=1,~k\neq m}}^{M}\mathcal{D}_k\tilde{x}_k^{t} + \mathcal{D}_mx_m - z^t\big\|^2, \nonumber\\&\hspace*{36pt} m=1,\cdots,M
%%%  \arg\min_{x_m}\hat{\mathcal{L}}_{\rho,t}(x_m,\tilde{x}_m^t,z^t,y^t).
  \nonumber\\
  &\xi_m^{t+1}:=\mathcal{N}(0,\sigma_{m,t+1}^2
  (\mathcal{D}_m^\top\mathcal{D}_m)^{-1}
  )\nonumber\\
  &\tilde{x}_m^{t+1}:= x_m^{t+1}+\xi_m^{t+1}\label{admmstepsdp}\\
  &z^{t+1}:=\underset{z}{\text{argmin}}\quad l(z)  - \langle y^t, z \rangle + \frac{\rho}{2} \big\|\sum_{m=1}^{M}\mathcal{D}_m\tilde{x}_m^{t+1} - z\big\|^2\nonumber\\
&y^{t+1}:=y^t + \rho\big(\sum_{m=1}^{M}\mathcal{D}_m\tilde{x}_m^{t+1} - z^{t+1}\big). \nonumber
\end{align}
%\textbf{where $x_{m,\textcolor{red}{\mathcal{D}_m}}^{t+1}$ is to emphasize that $x_{m}^{t+1}$
%is relied on data $\textcolor{red}{\mathcal{D}_m}$, however, in the following, with a slight abuse of notation, we will write $x_{m,\textcolor{red}{\mathcal{D}_m}}^{t+1}$ as $x_{m}^{t+1}$ in
%the case of no ambiguity.}
In the remaining part of this section, we demonstrate that (\ref{admmstepsdp}) guarantees $(\varepsilon, \delta)$~differential privacy with outputs $\{\mathcal{D}_m\tilde{x}_m^{t+1}\}_{t=0,1,\cdots,T-1}$ for some carefully selected $\sigma_{m,t+1}$. Beside Assumption~\ref{theo:assumptions_pri}, we introduce another set of assumptions widely used by the literature.
\begin{assume}\label{theo:assumptions_pri_added}
  \begin{enumerate}
    \item The feasible set $\{x,y\}$ and the dual variable $z$ are bounded; their $l_2$ norms have an upper bound $b_1$.\label{item:assum_5_pri}
    \item The regularizer $R_m(\cdot)$ is doubly differentiable
    with $|R_m^{\prime\prime}(\cdot)|\leq c_1$, where $c_1$ is a finite constant.\label{item:assum_6_pri}
    \item Each row of $\mathcal{D}_m$ is normalized and has an $l_2$ norm of 1.\label{item:assum_7_pri}
  \end{enumerate}
\end{assume}
Note that Assumption \ref{theo:assumptions_pri_added}.\ref{item:assum_5_pri} is adopted in \cite{sarwate2013signal} and \cite{wang2019global}. Assumption \ref{theo:assumptions_pri_added}.\ref{item:assum_6_pri} comes from \cite{zhang2016dynamic} and Assumption \ref{theo:assumptions_pri_added}.\ref{item:assum_7_pri} comes from \cite{zhang2016dynamic}
and \cite{sarwate2013signal}. As a typical method in differential privacy analysis, we first study the $l_2$ sensitivity of
$\mathcal{D}_mx_m^{t+1}$, which is defined by:
\begin{defi}
The $l_2$-norm sensitivity of $\mathcal{D}_mx_m^{t+1}$ is defined by:
  \begin{eqnarray*}
\Delta_{m,2}=\max_{\substack{\mathcal{D}_m,D_m^{\prime}\\
\|\mathcal{D}_m-D_m^{\prime}\|\leq1
}}\|\mathcal{D}_mx_{m,\mathcal{D}_m}^{t+1}
-\mathcal{D}_m^{\prime}x_{m,\mathcal{D}_m^{\prime}}^{t+1}\|.
  \end{eqnarray*}
  where $\mathcal{D}_m$ and $\mathcal{D}_m^{\prime}$ are two neighbouring datasets differing in 
  only one feature column, and 
  $x_{m,\mathcal{D}_m}^{t+1}$ is the $x_m^{t+1}$ derived from the first line of equation 
  (\ref{admmstepsdp}) under dataset $\mathcal{D}_m$.
\end{defi}
We have Lemma~\ref{lemma:privacy} to state the upper bound of the $l_2$-norm sensitivity of $\mathcal{D}_mx_m^{t+1}$.
\begin{lemma}
\label{lemma:privacy}
  Assume that Assumption~\ref{theo:assumptions_pri} and Assumption~\ref{theo:assumptions_pri_added} hold.
  %  the $l_2$-norm sensitivity of $x_m^{t+1}$ is defined by:
%  \begin{eqnarray*}
%  % \nonumber % Remove numbering (before each equation)
%\Delta_{m,2}=\max_{\substack{\mathcal{D}_m,D_m^{\prime}\\
%\|\mathcal{D}_m-D_m^{\prime}\|\leq1
%}}\|\mathcal{D}_mx_{m,\mathcal{D}_m}^{t+1}
%-\mathcal{D}_m^{\prime}x_{m,\mathcal{D}_m^{\prime}}^{t+1}\|
%  \end{eqnarray*}
Then the $l_2$-norm sensitivity of $\mathcal{D}_mx_{m,\mathcal{D}_m}^{t+1}$ is upper bounded by $\mathbb{C}=\frac{3}{d_m\rho}\left[\lambda c_1+(1+M\rho)b_1\right]$.
\end{lemma}

\begin{theorem}\label{theo:DP}
  Assume assumptions \ref{theo:assumptions_pri_added}.\ref{item:assum_1_pri}-\ref{theo:assumptions_pri_added}.\ref{item:assum_7_pri} hold and $\mathbb{C}$ is the 
  upper bound of $\Delta_{m,2}$. Let $\varepsilon\in(0,1]$ be an arbitrary constant and let
  $\mathcal{D}_m\xi_m^{t+1}$ be sampled from zero-mean Gaussian distribution with variance $\sigma_{m,t+1}^2$,
  where
  \begin{eqnarray*}
    \sigma_{m,t+1}=\frac{\sqrt{2\text{ln}(1.25/\delta)}\mathbb{C}}{\varepsilon}.
  \end{eqnarray*}
  Then each iteration guarantees $(\varepsilon,\delta)$-differential privacy. Specifically,
  for any neighboring datasets $\mathcal{D}_m$ and $\mathcal{D}_m^{\prime}$, for any output
  $\mathcal{D}_m\tilde{x}_{m,\mathcal{D}_m}^{t+1}$ and $\mathcal{D}_m^{\prime}\tilde{x}_{m,\mathcal{D}_m^{\prime}}^{t+1}$, the following inequality always holds:
  \begin{eqnarray*}
    P(\mathcal{D}_m\tilde{x}_{m,\mathcal{D}_m}^{t+1}|\mathcal{D}_m)\leq e^{\varepsilon}
    P(\mathcal{D}_m^{\prime}\tilde{x}_{m,\mathcal{D}_m^{\prime}}^{t+1}|\mathcal{D}_m^{\prime})
    +\delta.
  \end{eqnarray*}
\end{theorem}

With an application of the composition theory in \cite{dwork2014algorithmic}, we come to a result stating the overall privacy guarantee for the whole training procedure.
\begin{coro}
  For any $\delta^{\prime}>0$, the algorithm described in \eqref{admmstepsdp} satisfies $(\varepsilon^{\prime}, T\delta+\delta^{\prime})-$differential privacy within $T$ epochs of updates, where
  \begin{equation}
    \varepsilon^{\prime}=\sqrt{2T\text{ln}(1/\delta^{\prime})}\varepsilon+T\varepsilon(e^\varepsilon - 1).
  \end{equation}
\end{coro}

\section{Experiments}

\begin{figure*}[t]
    \centering
    % \vspace{-5mm}
    \subfigure[Loss vs. epoch]{
    \includegraphics[width=2.5in]{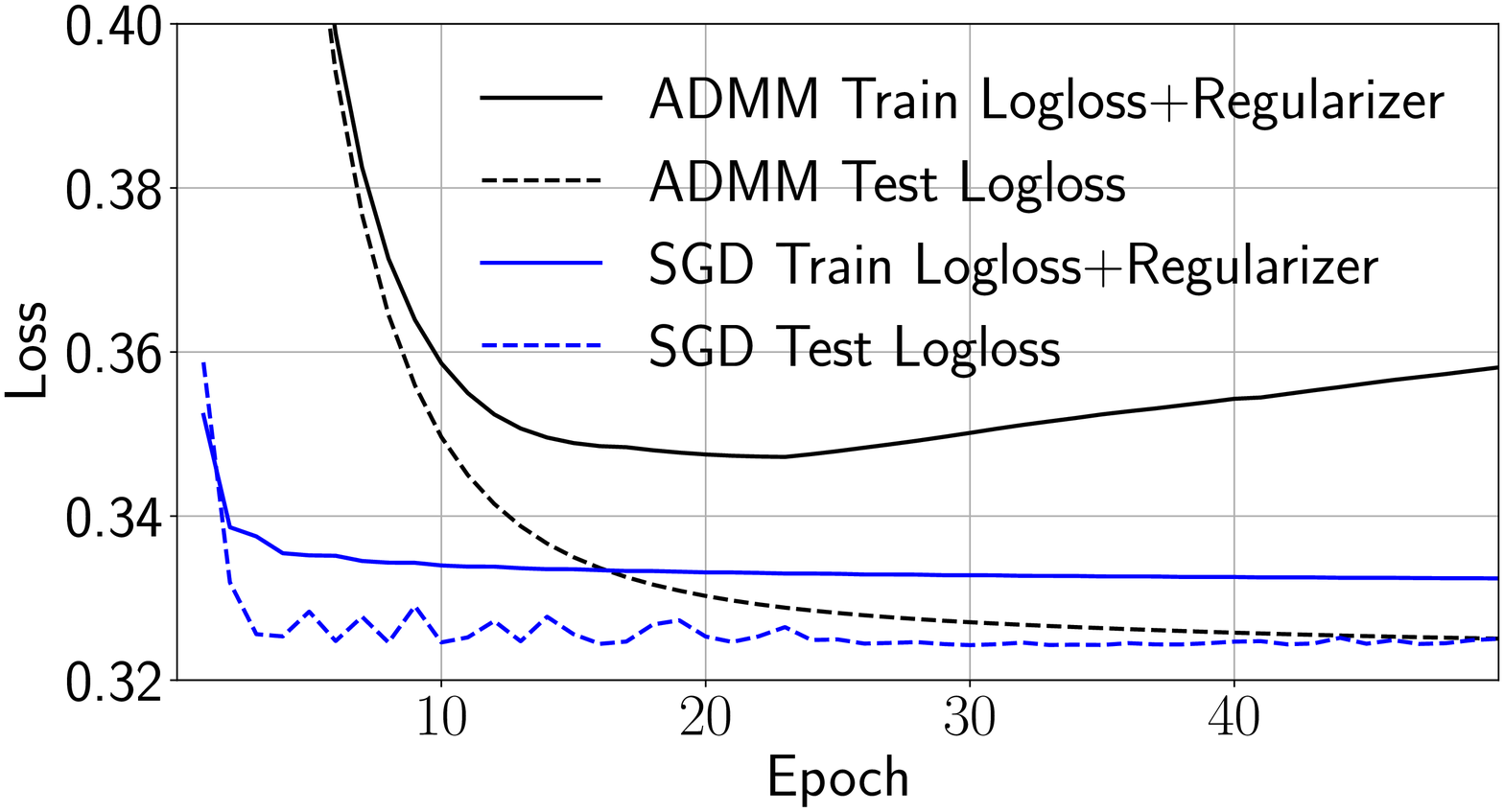}
    \label{fig:a9a_loss_vs_epoch}
    }
    % \hspace{-3.5mm}
    \subfigure[Test log loss under different noise levels]{
    \includegraphics[width=2.5in]{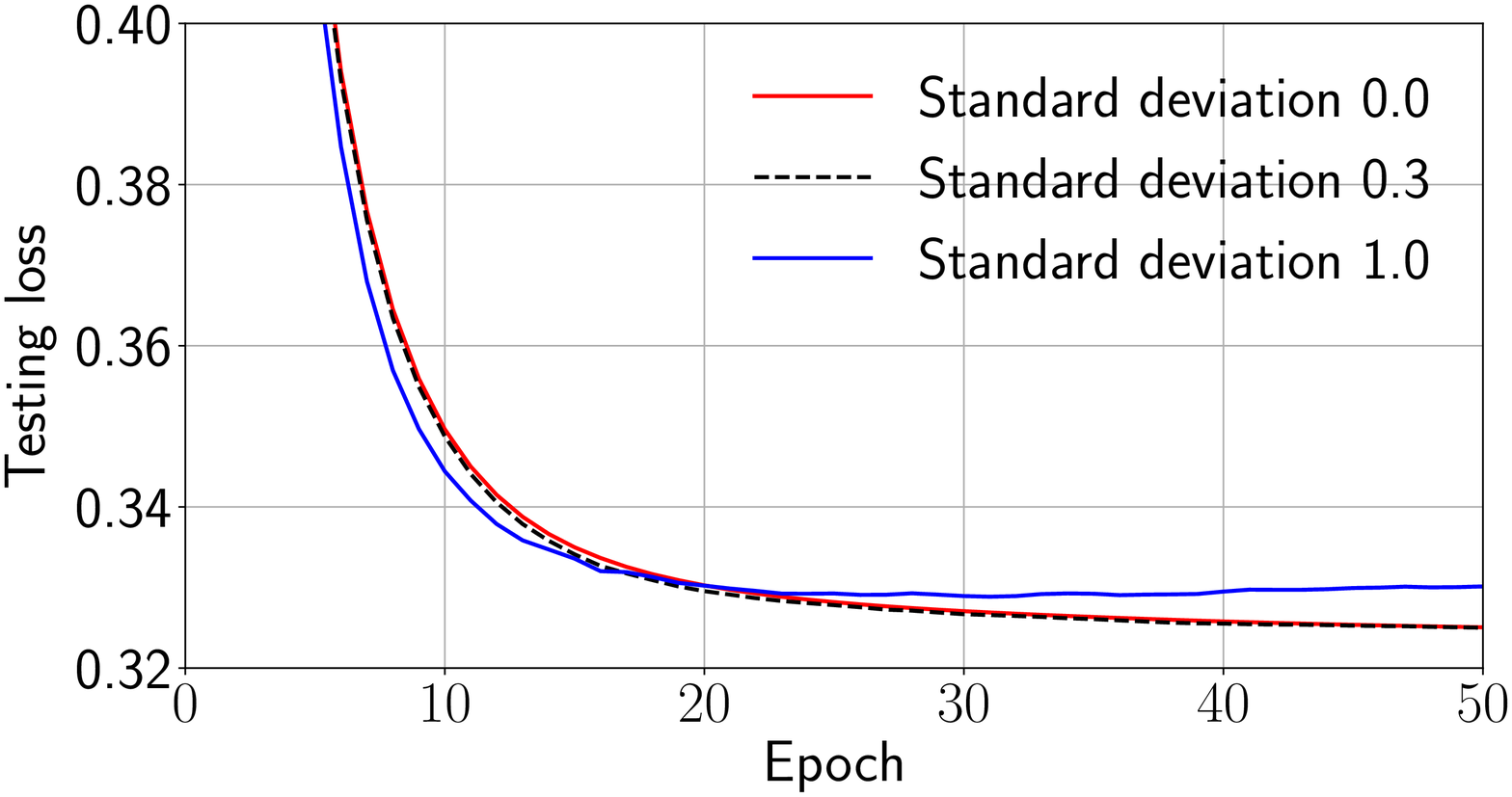}
    \label{fig:a9a_loss_vs_time}
    }
    \vspace{-3.5mm}
    \caption{Performance over the \emph{a9a} data set with 32561 training samples, 16281 testing samples and 123 features.}
    \label{fig:a9a_train}
    \vspace{-4mm}
\end{figure*}

\begin{figure*}[t]
    \centering
    % \vspace{-5mm}
    \subfigure[Loss vs. epoch]{
    \includegraphics[width=2.5in]{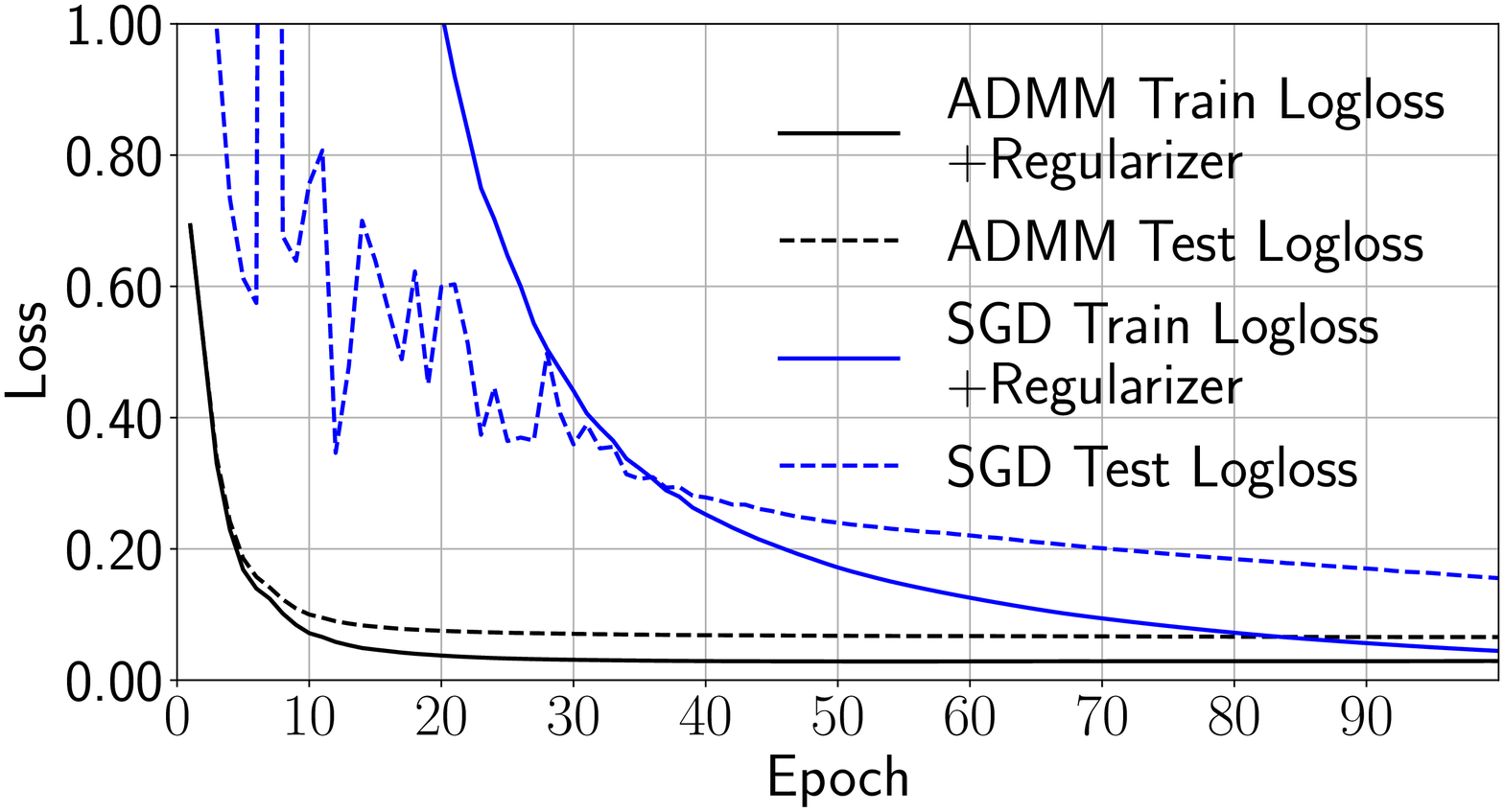}
    \label{fig:gisette_loss_vs_epoch}
    }
    % \hspace{-3.5mm}
    \subfigure[Test log loss under different noise levels]{
    \includegraphics[width=2.5in]{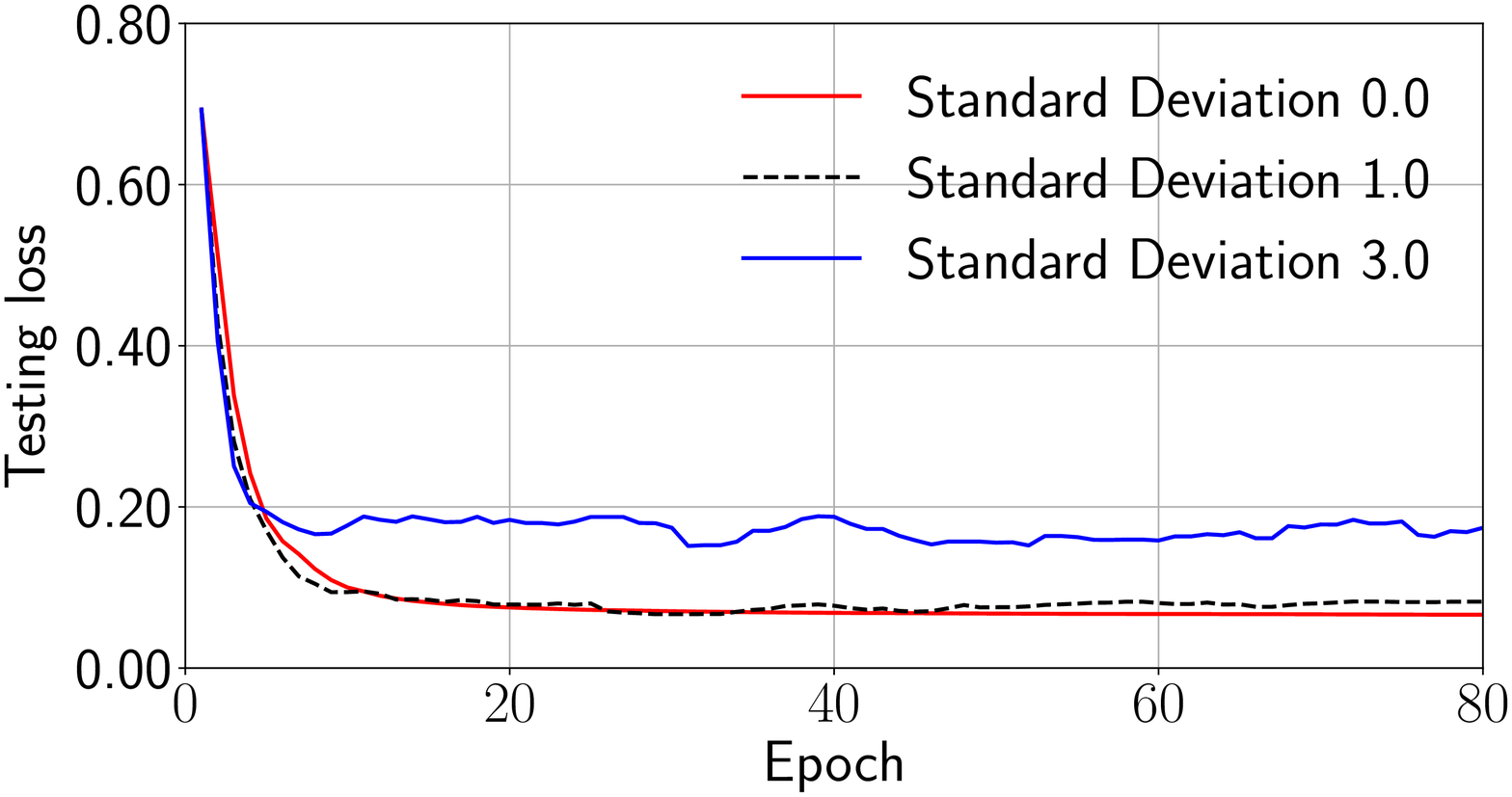}
    \label{fig:gisette_loss_vs_time}
    }
    \vspace{-3.5mm}
    \caption{Performance over the \emph{gisette} data set with 6000 training samples, 1000 testing samples and 5000 features.}
    \label{fig:gisette_train}
    \vspace{-4mm}
\end{figure*}

\begin{figure*}[t]
    \centering
    % \vspace{-5mm}
    \subfigure[\emph{a9a} data set]{
    \includegraphics[width=2.5in]{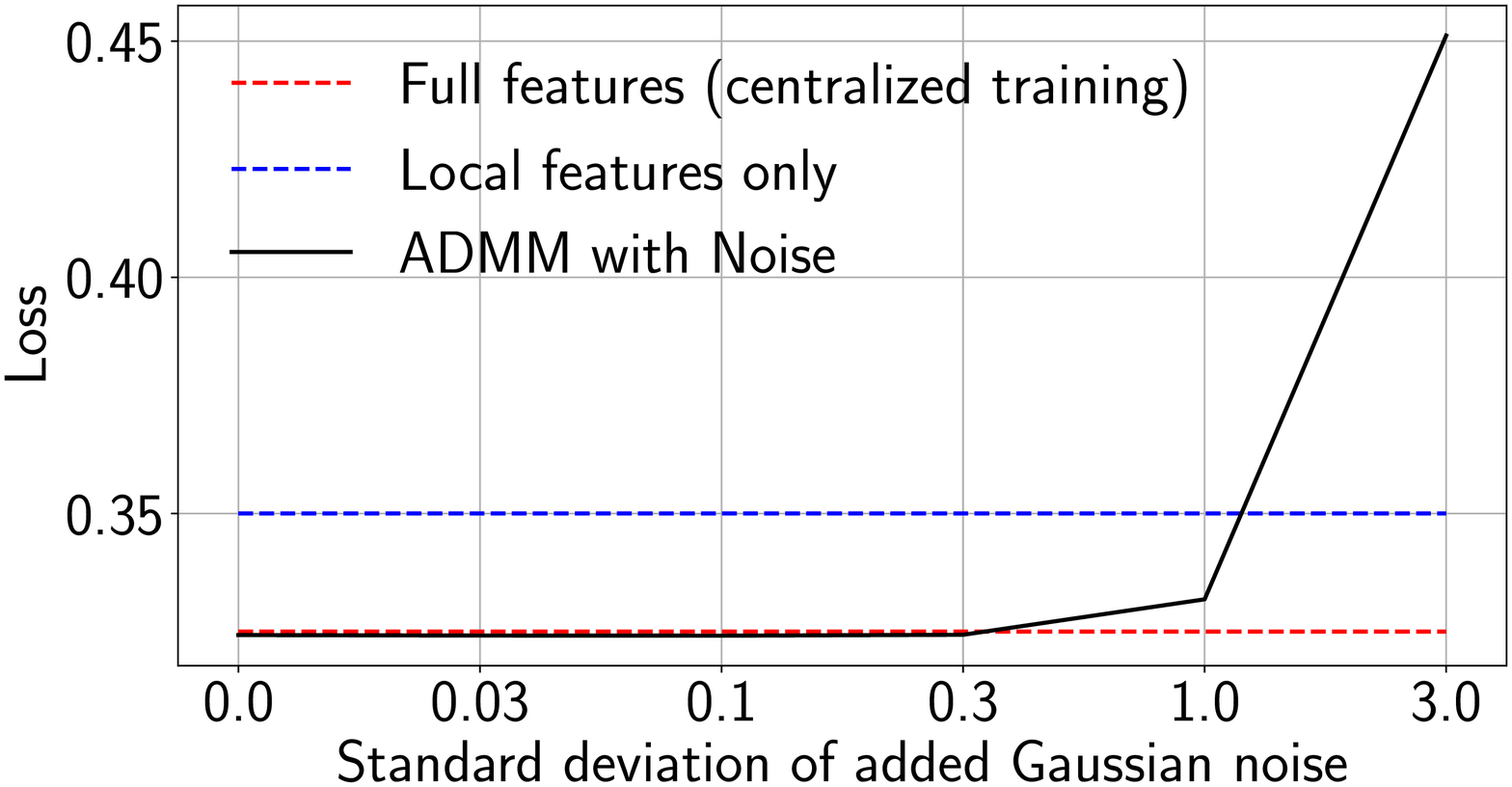}
    \label{fig:a9a_test_loss_vs_noise}
    }
    % \hspace{-3.5mm}
    \subfigure[\emph{gisette} data set]{
    \includegraphics[width=2.5in]{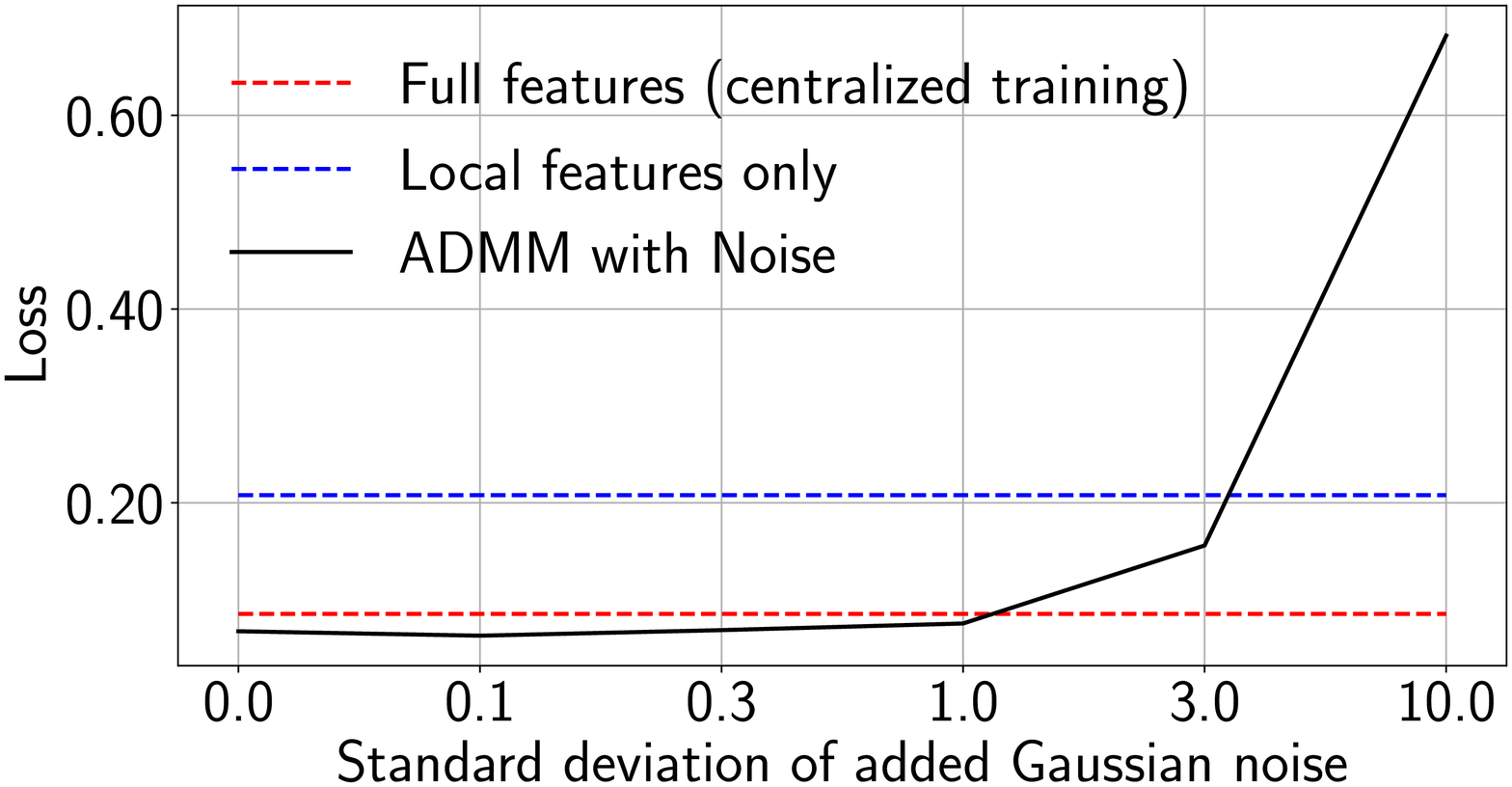}
    \label{fig:gisette_test_loss_vs_noise}
    }
    \vspace{-3.5mm}
    \caption{Test performance for ADMM under different levels of added noise.}
    \label{fig:loss_vs_noise}
     \vspace{-4mm}
\end{figure*}

We test our algorithm by training $l_2$-norm regularized logistic regression on two popular public datasets, namely, \emph{a9a} from UCI \cite{Dua:2019} and \emph{giette} \cite{guyon2005result}. We get the datasets from \cite{Liblinear:2019} so that we follow the same preprocessing procedure listed there. \emph{a9a} dataset is $4$ MB and contains 32561 training samples, 16281 testing samples and 123 features. We divide the dataset into two parts, with the first part containing the first 66 features and the second part remaining 57 features. The first part is regarded as the local party who wishes to improve its prediction model with the help of data from the other party. On the other hand, \emph{gisette} dataset is $297$ MB and contains 6000 training samples, 1000 testing samples and 5000 features. Similarly, we divide the features into 3 parts, the first 2000 features being the first part regarded as the local data, the next 2000 features being the second part, and the remaining 1000 as the third part. Note that \emph{a9a} is small in terms of the number of features and \emph{gisette} has a relatively higher dimensional feature space.

A prototype system is implemented in \emph{Python} to verify our proposed algorithm. Specifically, we use optimization library from \emph{scipy} to handle the optimization subproblems. We apply L-BFGS-B algorithm to do the $x$ update in \eqref{eq:pal_algo_x} and entry-wise optimization for $z$ in \eqref{eq:pal_algo_z}. We run the experiment on a machine equipped with Intel(R) Core(TM) i9-9900X CPU @ 3.50GHz and 128 GB of memory. 

We compare our algorithm against an SGD based algorithm proposed in \cite{hu2019fdml}.
We keep track of the training objective value (log loss plus the $l_2$ regularizer), the testing log loss for each epoch for different datasets and parameter settings. We also test our algorithm with different levels of Gaussian noise added. In the training procedure, we initialize the elements in $x$, $y$ and $z$ with $0$ while we initialize the parameter for the SGD-based algorithm with random numbers.

Fig.~\ref{fig:a9a_train} and Fig.~\ref{fig:gisette_train} show a typical trace of the training objective and testing log loss against epochs for \emph{a9a} and \emph{gisette}, respectively. On \emph{a9a}, the ADMM algorithm is slightly slower than the SGD based algorithm, while they reach the same testing log loss in the end. On \emph{gisette}, the SGD based algorithm converges slowly while the ADMM algorithm is efficient and robust. The testing log loss from the ADMM algorithm quickly converges to 0.08 after a few epochs, but the SGD based algorithm converges to only 0.1 with much more epochs. This shows that the ADMM algorithm is superior when the number of features is large. In fact, for each epoch, the $x$ update is a trivial quadratic program and can be efficiently solved numerically. The $z$ update contains optimization over computationally expensive functions, but for each sample, it is always an optimization over a single scalar so that it can be solved efficiently via scalar optimization and scales with the number of features.

Fig.~\ref{fig:loss_vs_noise} shows the testing loss for ADMM with different levels of Gaussian noise added. The other two baselines are the logistic regression model trained over all the features (in a centralized way) and that trained over only the local features in the first party. The baselines are trained with the built-in logistic regression function from \emph{sklearn} library. We can see that there is a significant performance boost if we employ more features to help training the model on Party 1. Interestingly, in Fig.~\ref{fig:gisette_test_loss_vs_noise}, the ADMM sharing has even better performance than the baseline trained with all features with \emph{sklearn}. It further shows that the ADMM sharing is better at datasets with a large number of features. 

Moreover, after applying moderate random perturbations, the proposed algorithm can still converge in a relatively small number of epochs, as Fig.~\ref{fig:a9a_loss_vs_time} and Fig.~\ref{fig:gisette_loss_vs_time} suggest, although too much noise may ruin the model.
%Thus, we can still achieve better performance than using local features only, although too much noise will ruin the model. 
Therefore, ADMM sharing algorithm under moderate perturbation can improve the local model and the privacy cost is well contained as the algorithm converges in a few epochs.

\section{Conclusion}

We study learning over distributed features (vertically partitioned data) where none of the parties shall share the local data. %The motivation is in contrast to the most existing literature on collaborative and distributed machine learning where the data samples (but not the features) are assumed to distribute and work in a data-parallel fashion. 
We propose the parallel ADMM sharing algorithm to solve this challenging problem where only intermediate values are shared, without even sharing model parameters. We have shown the convergence for convex and non-convex loss functions. To further protect the data privacy, we apply the differential privacy technique in the training procedure to derive a privacy guarantee within $T$ epochs. 
We implement a prototype system and evaluate the proposed algorithm on two representative datasets in risk minimization. The result shows that the ADMM sharing algorithm converges efficiently, especially on dataset with large number of features. Furthermore, the differentially private ADMM algorithm yields better prediction accuracy than model trained from only local features while ensuring a certain level of differential privacy guarantee. 

\newpage
\bibliographystyle{ACM-Reference-Format}
\bibliography{ms}

\newpage
\section{Supplementary Materials}
\subsection{Proof of Theorem~\ref{theo:convergence}}
To help theoretical analysis, we denote the objective functions in \eqref{eq:pal_algo_x} and \eqref{eq:pal_algo_z} as
\begin{align}
    & g_m(x_m) = \lambda R_m(x_m) + \langle y^t, \mathcal{D}_mx_m\rangle + \frac{\rho}{2}\big\|\sum_{\substack{k=1\\k\neq m}}^{M} \mathcal{D}_kx_k^t+ \mathcal{D}_mx_m - z^t\big\|^2,\nonumber\\
    & h(z) = l(z)  - \langle y^t, z \rangle + \frac{\rho}{2} \big\|\sum_{m=1}^{M}\mathcal{D}_mx_m^{t+1} - z\big\|^2,
\end{align}
correspondingly. We prove the following four lemmas to help prove the theorem. 
\begin{lemma}\label{lemma:y_diff_bound}
Under Assumption~\ref{theo:assumptions_pri}, we have
\[
    \nabla l(z^{t+1}) = y^{t+1},
\]
and
\[
    \|y^{t+1} - y^t\|^2 \le L^2\|z^{t+1} - z^{t}\|^2.
\]
\end{lemma}
{\bf Proof.} By the optimality in \eqref{eq:pal_algo_z}, we have
\begin{align}
    \nabla l(z^{t+1}) - y^t + \rho\big(z^{t+1} - \sum_{m=1}^{M} \mathcal{D}_mx_m^{t+1}\big) = 0.\nonumber
\end{align}
Combined with \eqref{eq:pal_algo_y}, we can get
\begin{align}
    \nabla l(z^{t+1}) = y^{t+1}.
\end{align}
Combined with Assumption~\ref{theo:assumptions_pri}.\ref{item:assum_1_pri}, we have
\begin{align}
\|y^{t+1} - y^{t}\|^2 = \|\nabla l(z^{t+1}) - \nabla l(z^{t})\|^2 \le L^2\|z^{t+1} - z^t\|^2.
\end{align}
\hfill$\square$

\begin{lemma}\label{lemma:algebra_1}
    We have 
    \begin{align}
        & \big(\|\sum_{m=1}^{M} x_m^{t+1} - z\|^2 - \|\sum_{m=1}^{M} x_m^{t} - z\|^2\big) - \sum_{m=1}^{M}\big(\|\sum_{\substack{k=1\\k\neq m}}^M x_k^{t} + x_m^{t+1} - z\|^2 - \|\sum_{m=1}^{M} x_m^{t} - z\|^2\big)\nonumber\\
        \le & \sum_{m=1}^{M} \|x_m^{t+1} - x_m^{t}\|^2.
    \end{align}
\end{lemma}
{\bf Proof.}
\begin{align}
    \text{LHS} = & \big(\sum_{m=1}^{M}(x_m^{t+1} + x_m^{t}) - 2z\big)\top(\sum_{m=1}^{M} x_m^{t+1} - \sum_{m=1}^{M} x_m^t) - \sum_{m=1}^{M} \big(\sum_{\substack{k=1\\k\neq m}}^M2x_k^t + x_m^t + x_m^{t+1} -2z\big)\top(x_m^{t+1} - x_m^t)\nonumber\\
    = & - \sum_{m=1}^{M} \sum_{\substack{k=1\\k\neq m}}^M(x_k^{t+1} - x_k^{t})\top(x_m^{t+1} - x_m^t)\nonumber\\
    = & - \|\sum_{m=1}^{M} (x_m^{t+1} - x_m^{t})\|^2 + \sum_{m=1}^{M} \|x_m^{t+1} - x_m^{t}\|^2\nonumber\\
    \le & \sum_{m=1}^{M} \|x_m^{t+1} - x_m^{t}\|^2.\nonumber
\end{align}
\hfill$\square$

\begin{lemma}\label{lemma:L_iter_diff}
Suppose Assumption~\ref{theo:assumptions_pri} holds. We have
\begin{align}
    & \mathcal{L}(\{x_m^{t+1}\},z^{t+1};y^{t+1}) - \mathcal{L}(\{x_m^{t}\},z^{t};y^{t})\nonumber\\
    & \le \sum_{m=1}^{M} -\left(\frac{\gamma_m(\rho)}{2}-\sigma_{\text{max}}(\mathcal{D}_m^\top \mathcal{D}_m)\right)\|x_m^{t+1} - x_m^t\|^2 - \left(\frac{\gamma(\rho)}{2} - \frac{L^2}{\rho}\right)\|z^{t+1} - z^t\|^2. \nonumber
\end{align}
\end{lemma}
{\bf Proof.} The LFH can be decomposed into two parts as
\begin{align}
    & \mathcal{L}(\{x_m^{t+1}\},z^{t+1};y^{t+1}) - \mathcal{L}(\{x_m^{t}\},z^{t};y^{t})\nonumber\\
    = & \big(\mathcal{L}(\{x_m^{t+1}\},z^{t+1};y^{t+1}) - \mathcal{L}(\{x_m^{t+1}\},z^{t+1};y^{t})\big)\nonumber\\
    & + \big(\mathcal{L}(\{x_m^{t+1}\},z^{t+1};y^{t}) - \mathcal{L}(\{x_m^{t}\},z^{t};y^{t})\big). \label{eq:proof_diff_L_0_pal}
\end{align}
For the first term, we have
\begin{align}
    & \mathcal{L}(\{x_j^{t+1}\},z^{t+1};y^{t+1}) - \mathcal{L}(\{x_j^{t+1}\},z^{t+1};y^{t})\nonumber\\
    = & \langle y^{t+1} - y^t , \sum_jD_jx_j^{t+1} - z^{t+1}\rangle\nonumber\\
    = & \frac{1}{\rho}\|y^{t+1} - y^t\|^2\quad\text{(by \eqref{eq:pal_algo_y})}\nonumber\\
    = & \frac{L^2}{\rho}\|z^{t+1} - z^t\|^2\quad\text{(by Lemma~\ref{lemma:y_diff_bound})}\label{eq:proof_diff_L_1_pal}.
\end{align}
For the second term, we have
\begin{align}
    & \mathcal{L}(\{x_m^{t+1}\},z^{t+1};y^{t}) - \mathcal{L}(\{x_m^{t}\},z^{t};y^{t})\nonumber\\
    = & \mathcal{L}(\{x_m^{t+1}\},z^{t+1};y^{t}) - \mathcal{L}(\{x_m^{t}\},z^{t+1};y^{t}) + \mathcal{L}(\{x_m^{t}\},z^{t+1};y^{t}) - \mathcal{L}(\{x_m^{t}\},z^{t};y^{t})\nonumber\\
    \le & \bigg(\big(\lambda \sum_{m=1}^{M}R_m(x_m^{t+1}) + \langle y^t, \sum_{m=1}^{M}\mathcal{D}_mx_m^{t+1}\rangle + \frac{\rho}{2}\big\|\sum_{k=1}^M \mathcal{D}_kx_k^{t+1} - z^{t+1}\big\|^2\big)\nonumber\\
    & - \big(\lambda \sum_{m=1}^{M}R_m(x_m^{t}) + \langle y^t, \sum_{m=1}^{M}\mathcal{D}_mx_m^{t}\rangle + \frac{\rho}{2}\big\|\sum_{k=1}^M \mathcal{D}_kx_k^{t} - z^{t+1}\big\|^2\big)\bigg)\nonumber\\
    & + \bigg(\big(l(z^{t+1}) -\langle y^t, z^{t+1}\rangle + \frac{\rho}{2} \big\|\sum_{m=1}^{M} \mathcal{D}_mx_m^{t+1} - z^{t+1}\big\|^2\big)-\big(l(z^{t}) -\langle y^t, z^{t}\rangle+ \frac{\rho}{2} \big\|\sum_{m=1}^{M} \mathcal{D}_mx_m^{t+1} - z^{t}\big\|^2\big)\bigg)\nonumber\\
    \le & \sum_{m=1}^{M} \bigg(\big(\lambda R_m(x_m^{t+1}) + \langle y^t, \mathcal{D}_mx_m^{t+1}\rangle + \frac{\rho}{2}\big\|\sum_{\substack{k=1\\k\neq m}}^M\mathcal{D}_kx_k^{t} + \mathcal{D}_mx_m^{t+1} - z^{t+1}\big\|^2\big)\nonumber\\
    & - \big(\lambda R_m(x_m^{t}) + \langle y^t, \mathcal{D}_mx_m^{t}\rangle + \frac{\rho}{2}\big\|\sum_{k=1}^{M}\mathcal{D}_kx_k^{t} - z^{t+1}\big\|^2\big)\bigg)+\sum_{m=1}^{M} \|\mathcal{D}_m(x_m^{t+1} - x_m^t)\|^2\nonumber\\
    & + \bigg(\big(l(z^{t+1}) -\langle y^t, z^{t+1}\rangle + \frac{\rho}{2} \big\|\sum_{m=1}^{M} \mathcal{D}_mx_m^{t+1} - z^{t+1}\big\|^2\big)\nonumber\\
    & -\big(l(z^{t}) -\langle y^t, z^{t}\rangle+ \frac{\rho}{2} \big\|\sum_{m=1}^{M} \mathcal{D}_mx_m^{t+1} - z^{t}\big\|^2\big)\bigg)\quad\text{(by Lemma~\ref{lemma:algebra_1})}\nonumber\\
    = & \sum_{m=1}^{M} \big(g_m(x_m^{t+1}) - g_m(x_m^t)\big) + (h(z^{t+1}) - h(z^T)))+\sum_{m=1}^{M} \|\mathcal{D}_m(x_m^{t+1} - x_m^t)\|^2\nonumber\\
    \le & \sum_{m=1}^{M} \big(\langle \nabla g_m(x_m^{t+1}), x_m^{t+1}-x_m^t\rangle - \frac{\gamma_m(\rho)}{2}\|x_m^{t+1}-x_m^t\|^2\big) + \langle\nabla h(z^{t+1}), z^{t+1} - z^t\rangle - \frac{\gamma(\rho)}{2}\|z^{t+1} - z^t\|^2\nonumber\\
    & +\sum_{m=1}^{M} \|\mathcal{D}_m(x_m^{t+1} - x_m^t)\|^2\quad\text{(by strongly convexity from Assumption~\ref{theo:assumptions_pri}.\ref{item:assum_2_pri})}\nonumber
\end{align}
\begin{align}
    \le & -\sum_{m=1}^{M}\frac{\gamma_m(\rho)}{2}\|x_m^{t+1}-x_m^t\|^2 - \frac{\gamma(\rho)}{2}\|z^{t+1} - z^t\|^2+\sum_{m=1}^{M}\|\mathcal{D}_m(x_m^{t+1} - x_m^t)\|^2\nonumber\\
    & \text{(by optimality condition for subproblem in \eqref{eq:pal_algo_x} and \eqref{eq:pal_algo_z})}\nonumber\\
    \le & \sum_{m=1}^{M} -\left(\frac{\gamma_m(\rho)}{2}-\sigma_{\text{max}}(\mathcal{D}_m^\top \mathcal{D}_m)\right)\|x_m^{t+1} - x_m^t\|^2 - \frac{\gamma(\rho)}{2}\|z^{t+1} - z^t\|^2.\label{eq:proof_diff_L_2_pal}
\end{align}
Note that we have abused the notation $\nabla g_m(x_m)$ and denote it as the subgradient when $g$ is non-smooth but convex.
Combining \eqref{eq:proof_diff_L_0_pal}, \eqref{eq:proof_diff_L_1_pal} and \eqref{eq:proof_diff_L_2_pal}, the lemma is proved.\hfill$\square$

\begin{lemma}\label{lemma:L_lower_bound}
    Suppose Assumption~\ref{theo:assumptions_pri} holds. Then the following limit exists and is bounded from below:
    \begin{align}
        \underset{t\rightarrow\infty}{\text{lim}} \mathcal{L}(\{x^{t+1}\}, z^{t+1}; y^{t+1}).
    \end{align}
\end{lemma}
{\bf Proof.}
\begin{align}
    & \mathcal{L}(\{x^{t+1}\}, z^{t+1}; y^{t+1} )\nonumber\\
    = & l(z^{t+1}) + \lambda\sum_{m=1}^{M} R_m(x_m^{t+1}) + \langle y^{t+1}, \sum_{m=1}^{M}\mathcal{D}_m x_m^{t+1} - z^{t+1}\rangle + \frac{\rho}{2}\|\mathcal{D}_m x_m^{t+1} - z^{t+1}\|^2\nonumber\\
    = &\lambda\sum_{m=1}^{M} R_m(x_m^{t+1}) + l(z^{t+1}) + \langle \nabla l(z^{t+1}), \sum_{m=1}^{M}\mathcal{D}_m x_m^{t+1} - z^{t+1}\rangle + \frac{\rho}{2}\|\mathcal{D}_m x_m^{t+1} - z^{t+1}\|^2\quad\text{(by Lemma~\ref{lemma:y_diff_bound})}\nonumber\\
    \ge & \lambda\sum_{m=1}^{M} R_m(x_m^{t+1}) + l(\sum_{m=1}^{M}\mathcal{D}_m x_m^{t+1}) + \frac{\rho-L}{2}\|\mathcal{D}_m x_m^{t+1} - z^{t+1}\|^2.\quad%\text{(from Assumption~\ref{theo:assumptions_pal}.\ref{item:assum_1_pal})}
\end{align}
Combined with Assumption~\ref{theo:assumptions_pri}.\ref{item:assum_3_pri}, $\mathcal{L}(\{x^{t+1}\}, z^{t+1}; y^{t+1})$ is lower bounded. Furthermore, by Assumption~\ref{theo:assumptions_pri}.\ref{item:assum_2_pri} and Lemma~\ref{lemma:L_iter_diff}, $\mathcal{L}(\{x^{t+1}\}, z^{t+1}; y^{t+1})$ is decreasing. These complete the proof. \hfill$\square$

Now we are ready for the proof.

{\bf Part 1. } By Assumption~\ref{theo:assumptions_pri}.\ref{item:assum_2_pri}, Lemma~\ref{lemma:L_iter_diff} and Lemma~\ref{lemma:L_lower_bound}, we have
\begin{align}
    & \|x_m^{t+1} - x_m^t\|\rightarrow 0\quad\forall m=1,2,\ldots, M,\nonumber\\
    & \|z^{t+1} - z^t\|\rightarrow 0.\nonumber
\end{align}
Combined with Lemma~\ref{lemma:y_diff_bound}, we have
\begin{align}
    \|y^{t+1}-y^t\|^2\rightarrow 0.\nonumber
\end{align}
Combined with~\eqref{eq:pal_algo_y}, we complete the proof for Part 1.

{\bf Part 2.} Due to the fact that $\|y^{t+1}-y^t\|^2\rightarrow 0$, by taking limit in \eqref{eq:pal_algo_y}, we can get \eqref{eq:cond_primal}.

At each iteration $t+1$, by the optimality of the subproblem in \eqref{eq:pal_algo_z}, we have
\begin{align}
    \nabla l(z^{t+1}) - y^t + \rho (z^{t+1} - \sum_{m=1}^{M}\mathcal{D}_mx_m^{t+1}) = 0.
\end{align}
Combined with \eqref{eq:cond_primal} and taking the limit, we get \eqref{eq:cond_dual}.

Similarly, by the optimality of the subproblem in \eqref{eq:pal_algo_x}, for $\forall m\in\mathcal{M}$ there exists $\eta_m^{t+1}\in\partial R_m(x_m^{t+1})$, such that
\begin{align}
    \big\langle x_m-x_m^{t+1}, \lambda \eta_m^{t+1} + \mathcal{D}_m^Ty^t + \rho \mathcal{D}_m^T\big(\sum_{\substack{k=1\\k\le m}}^M \mathcal{D}_kx_k^{t+1} + \sum_{\substack{k=1\\k>j}}^M \mathcal{D}_kx_k^{t} - z^{t}\big)\big\rangle \ge 0\quad \forall x_m\in X_m.
\end{align}
Since $R_m$ is convex, we have
\begin{align}
    \lambda R_m(x_m) - \lambda R_m(x_m^{t+1}) + \big\langle x-x_m^{t+1}, \mathcal{D}_m^\top y^t + \rho\big(\sum_{\substack{k=1\\k\le m}}^M \mathcal{D}_kx_k^{t+1} + \sum_{\substack{k=1\\k>j}}^M \mathcal{D}_kx_k^{t} - z^{t}\big)^T \mathcal{D}_j\big\rangle \ge 0\quad \forall x_m\in X_m.
\end{align}
Combined with \eqref{eq:cond_primal} and the fact $\|x_m^{t+1} - x_m^{t}\|\rightarrow 0$, by taking the limit, we get
\begin{align}
    \lambda R_m(x_m) - \lambda R_m(x_m^*) + \big\langle x-x_m^{*}, \mathcal{D}_m^\top y^* \big\rangle \ge 0\quad \forall x_m\in X_m, \forall m,
\end{align}
which is equivalent to
\begin{align}
    \lambda R_m(x) + \big\langle y^*, \mathcal{D}_mx \big\rangle- \lambda R_m(x_m^*) - \big\langle y^*, \mathcal{D}_mx_m^{*} \big\rangle \ge 0\quad \forall x\in X_m, \forall m.
\end{align}
And we can get the result in \eqref{eq:cond_x_opt_conv}. \\
When $m\not\in\mathcal{M}$, we have
\begin{align}
    \big\langle x_m-x_m^{t+1}, \lambda \nabla R_m(x_m^{t+1}) + \mathcal{D}_m^\top y^t + \rho\big(\sum_{\substack{k=1\\k\le m}}^M \mathcal{D}_kx_k^{t+1} + \sum_{\substack{k=1\\k>j}}^M \mathcal{D}_kx_k^{t} - z^{t}\big)^T \mathcal{D}_m\big\rangle \ge 0\quad \forall x_m\in X_m.
\end{align}
Taking the limit and we can get \eqref{eq:cond_x_opt_nonconv}.

{\bf Part 3.} We first show that there exists a limit point for each of the sequences $\{x_m^t\}$, $\{z^t\}$ and $\{y^t\}$. Since $X_m, \forall m$ is compact, $\{x_m^t\}$ must have a limit point. With Theorem~\ref{theo:convergence}.\ref{item:primal_cond_limit}, we can get that $\{z^t\}$ is also compact and has a limit point. Furthermore, with Lemma~\ref{lemma:y_diff_bound}, we can get $\{y^t\}$ is also compact and has a limit point.

We prove Part 3 by contradiction. Since $\{x_m^t\}$, $\{z^t\}$ and $\{y^t\}$ lie in some compact set, there exists a subsequence $\{x_m^{t_k}\}$, $\{z^{t_k}\}$ and $\{y^{t_k}\}$, such that
\begin{align}
    (\{x_m^{t_k}\}, z^{t_k}; y^{t_k})\rightarrow (\{\hat{x}_m\}, \hat{z}; \hat{y}), \label{eq:subseq_limit}
\end{align}
where $(\{\hat{x}_m\}, \hat{z}; \hat{y})$ is some limit point and by part 2, we have $(\{\hat{x}_m\}, \hat{z}; \hat{y})\in Z^*$. Suppose that $\{\{x_m^t\}, z^t; y^t\}$ does not converge to $Z^*$, since $(\{x_m^{t_k}\}, z^{t_k}; y^{t_k})$ is a subsequence of it, there exists some $\gamma>0$, such that
\begin{align}
    \underset{k\rightarrow\infty}{\lim}\quad\text{dist}\big((\{x_m^{t_k}\}, z^{t_k}; y^{t_k});Z^*\big)=\gamma>0.\label{eq:suppose}
\end{align}
From \eqref{eq:subseq_limit}, there exists some $J(\gamma)>0$, such that
\begin{align}
        \|(\{x_m^{t_k}\}, z^{t_k}; y^{t_k})- (\{\hat{x}_m\}, \hat{z}; \hat{y})\|\le\frac{\gamma}{2}, \quad\forall k\ge J(\gamma).
\end{align}
Since $(\{\hat{x}_m\}, \hat{z}; \hat{y})\in Z^*$, we have
\begin{align}
    \text{dist}\big((\{x_m^{t_k}\}, z^{t_k}; y^{t_k});Z^*\big) \le \text{dist}\big((\{x_m^{t_k}\}, z^{t_k}; y^{t_k});  (\{\hat{x}_m\}, \hat{z}; \hat{y})\big).
\end{align}
From the above two inequalities, we must have
\begin{align}
    \text{dist}\big((\{x_m^{t_k}\}, z^{t_k}; y^{t_k});Z^*\big)\le\frac{\gamma}{2},\quad\forall k\ge J(\gamma),
\end{align}
which contradicts to \eqref{eq:suppose}, completing the proof.
\hfill$\square$

\subsection{Proof of Theorem~\ref{theo:iter_complexity}}
We first show an upper bound for $V^t$.

1. Bound for $\tilde{\nabla}_{x_m} \mathcal{L}(\{x_m^t\}, z^t; y^t)$. When $m\in\mathcal{M}$, from the optimality condition in \eqref{eq:pal_algo_x}, we have
\begin{align}
    0 \in \lambda\partial_{x_m}R_j(x_m^{t+1}) + \mathcal{D}^\top y^t + \rho \mathcal{D}_j^\top \big(\sum_{\substack{k=1\\k\neq j}}^M \mathcal{D}_kx_k^{t} + \mathcal{D}_mx_m^{t+1} - z^{t}\big).\nonumber
\end{align}
By some rearrangement, we have
\begin{align}
    \big(x_m^{t+1} - \mathcal{D}^\top y^t - \rho \mathcal{D}_m^\top \big(\sum_{\substack{k=1\\k\neq j}}^M \mathcal{D}_kx_k^{t}+ \mathcal{D}_mx_m^{t+1} - z^{t}\big)\big) - x_m^{t+1} \in \lambda\partial_{x_m}R_m(x_m^{t+1}),\nonumber
\end{align}
which is equivalent to
\begin{align}
    x_m^{t+1} = \text{prox}_{\lambda R_m}\big[x_m^{t+1} - \mathcal{D}^\top y^t - \rho \mathcal{D}_m^\top\big(\sum_{\substack{k=1\\k\neq j}}^M \mathcal{D}_kx_k^{t}+ \mathcal{D}_mx_m^{t+1} - z^{t}\big)\big].
\end{align}
Therefore,
\begin{align}
    & \|x_m^t - \text{prox}_{\lambda R_m} \big[x_m^t-\nabla_{x_m}\big(\mathcal{L}(\{x^t_m\}, z^t; y^t) - \lambda\sum_{m=1}^{M} R_m(x_m^t)\big)\big]\|\nonumber\\
    = & \big\|x_m^t - x_m^{t+1} + x_m^{t+1} - \text{prox}_{\lambda R_m}\big[x_m^{t} - \mathcal{D}^\top y^t - \rho \mathcal{D}_m^\top \big(\sum_{k=1}^M \mathcal{D}_kx_k^{t} - z^{t}\big)\big]\big\|\nonumber\\
    \le & \|x_m^t - x_m^{t+1}\| + \big\|  \text{prox}_{\lambda R_m}\big[x_m^{t+1} - \mathcal{D}^\top y^t - \rho \mathcal{D}_m^T\big( \sum_{\substack{k=1\\k\neq m}}^M \mathcal{D}_kx_k^{t}+ \mathcal{D}_mx_m^{t+1} - z^{t}\big)\big]\nonumber\\
    &  - \text{prox}_{\lambda R_m}\big[x_m^{t} - \mathcal{D}^\top y^t - \rho \mathcal{D}_m^T\big(\sum_{k=1}^M \mathcal{D}_kx_k^{t} - z^{t}\big)\big]\big\|\nonumber\\
    & \le 2\|x_m^t - x_m^{t+1}\| + \rho\|\mathcal{D}_m^\top \mathcal{D}_m(x_m^{t+1} - x_m^t)\|. \label{eq:proof_pal_bound_V_1}
\end{align}
When $m\not\in\mathcal{M}$, similarly, we have
\begin{align}
    \lambda\nabla_{x_m}R_m(x_m^{t+1}) + \mathcal{D}^\top y^t + \rho \mathcal{D}_m^\top \big(\sum_{\substack{k=1\\k\neq j}}^M \mathcal{D}_kx_k^{t}+ \mathcal{D}_mx_m^{t+1} - z^{t}\big) = 0.
\end{align}
Therefore,
\begin{align}
    & \|\nabla_{x_m} \mathcal{L}(\{x_m^t\}, z^t; y^t)\|\nonumber\\
    = & \|\lambda\nabla_{x_m}R_m(x_m^{t}) + \mathcal{D}^\top y^t + \rho \mathcal{D}_m^\top \big( \sum_{k=1}^M \mathcal{D}_kx_k^{t} - z^{t}\big)\|\nonumber\\
    = &  \|\lambda\nabla_{x_m}R_m(x_m^{t}) + \mathcal{D}^\top y^t + \rho \mathcal{D}_m^T\big( \sum_{k=1}^M \mathcal{D}_kx_k^{t} - z^{t}\big)\nonumber\\
     & - \big(\lambda\nabla_{x_m}R_m(x_m^{t+1}) + \mathcal{D}^\top y^t + \rho \mathcal{D}_m^T\big(\sum_{\substack{k=1\\k\neq j}}^M \mathcal{D}_kx_k^{t}+ \mathcal{D}_mx_m^{t+1} - z^{t}\big)\big)\|\nonumber\\
    \le &  \lambda\|\nabla_{x_m}R_m(x_m^{t}) - \nabla_{x_m}R_m(x_m^{t+1}) \| + \rho\|\mathcal{D}_m^\top \mathcal{D}_m(x_m^{t+1} - x_m^t)\|\nonumber\\
    \le & L_m\| x_m^{t+1} - x_m^t \|  + \rho\|\mathcal{D}_m^\top \mathcal{D}_m(x_m^{t+1} - x_m^t)\|. \quad\text{(by Assumption~\ref{theo:assumptions_pri}.\ref{item:assum_4_pri})}\label{eq:proof_pal_bound_V_2}
\end{align}

2. Bound for $\|\nabla_{z} \mathcal{L}(\{x_m^t\}, z^t; y^t)\|$. By optimality condition in \eqref{eq:pal_algo_z}, we have
\begin{align}
    \nabla l(z^{t+1}) - y^t + \rho\big(z^{t+1} - \sum_{m=1}^M \mathcal{D}_m x_m^{t+1}\big) = 0.\nonumber
\end{align}
Therefore
\begin{align}
        & \|\nabla_{z} \mathcal{L}(\{x_m^t\}, z^t; y^t)\|\nonumber\\
        = & \| l(z^{t}) - y^t + \rho\big(z^{t} - \sum_{m=1}^M \mathcal{D}_m x_m^{t}\big) \|\nonumber\\
        = & \|l(z^{t}) - y^t + \rho\big(z^{t} - \sum_{m=1}^M \mathcal{D}_m x_m^{t}\big) - \big(l(z^{t+1}) - y^t + \rho\big(z^{t+1} - \sum_{m=1}^M \mathcal{D}_m x_m^{t+1}\big)\big)\|\nonumber\\
        \le & (L+\rho)\|z^{t+1} - z^{t}\| + \rho\sum_{m=1}^M\|\mathcal{D}_m(x_m^{t+1}-x_m^{t})\|.\label{eq:proof_pal_bound_V_3}
\end{align}

3. Bound for $\|\sum_{m=1}^M \mathcal{D}_mx_m^t - z^t\|$. According to Lemma~\ref{lemma:y_diff_bound}, we have
\begin{align}
     \|\sum_{m=1}^M \mathcal{D}_mx_m^t - z^t\| = \frac{1}{\rho}\|y^{t+1}-y^t\| \le \frac{L}{\rho}\|z^{t+1}-z^t\|. \label{eq:proof_pal_bound_V_4}
\end{align}

Combining \eqref{eq:proof_pal_bound_V_1}, \eqref{eq:proof_pal_bound_V_2}, \eqref{eq:proof_pal_bound_V_3} and \eqref{eq:proof_pal_bound_V_4}, we can conclude that there exists some $C_1>0$, such that
\begin{align}
    V^t \le C_1(\|z^{t+1}-z^t\|^2 + \sum_{m=1}^M\|x_m^{t+1}-x_m^t\|^2), \label{eq:proof_pal_bound_V_5}
\end{align}

By Lemma~\ref{lemma:L_iter_diff}, there exists some constant $C_2 = \text{min}\{\sum_{m=1}^M\frac{\gamma_m(\rho)}{2}, \frac{\gamma(\rho)}{2} - \frac{L^2}{\rho}\}$, such that
\begin{align}
    & \mathcal{L}(\{x_m^t\}, z^t; y^t) - \mathcal{L}(\{x_m^{t+1}\}, z^{t+1}; y^{t+1})\nonumber\\
    \ge & C_2 (\|z^{t+1}-z^t\|^2 + \sum_{m=1}^M\|x_m^{t+1}-x_m^t\|^2). \label{eq:proof_pal_bound_V_6}
\end{align}

By \eqref{eq:proof_pal_bound_V_5} and \eqref{eq:proof_pal_bound_V_6}, we have
\begin{align}
    V^t\le \frac{C_1}{C_2}\mathcal{L}(\{x_m^t\}, z^t; y^t) - \mathcal{L}(\{x_m^{t+1}\}, z^{t+1}; y^{t+1}).
\end{align}
Taking the sum over $t=1,\ldots, T$, we have
\begin{align}
    \sum_{t=1}^T V^t \le & \frac{C_1}{C_2} \mathcal{L}(\{x^1\}, z^1; y^1) - \mathcal{L}(\{x^{t+1}\}, z^{t+1}; y^{t+1})\nonumber\\
    \le & \frac{C_1}{C_2} (\mathcal{L}(\{x^1\}, z^1; y^1) - \underline{f}).
\end{align}
By the definition of $T(\epsilon)$, we have
\begin{align}
    T(\epsilon)\epsilon \le \frac{C_1}{C_2} (\mathcal{L}(\{x^1\}, z^1; y^1)  - \underline{f}).
\end{align}
By taking $C = \frac{C_1}{C_2}$, we complete the proof. \hfill$\square$

\subsection{Proof of Lemma~\ref{lemma:privacy}}
From the optimality condition of the $x$ update procedure in (\ref{admmstepsdp}), we can get

\begin{eqnarray*}
  \mathcal{D}_mx_{m,\mathcal{D}_m}^{t+1}&=&-\mathcal{D}_m(\rho \mathcal{D}_m^\top \mathcal{D}_m)^{-1}
  \left[
  \lambda R_m^{\prime}(x_{m,\mathcal{D}_m}^{t+1})+\mathcal{D}_m^\top y^t+\rho\mathcal{D}_m^\top(\sum_{\substack
{k=1\\k\neq m}}^{M}\mathcal{D}_k\tilde{x}_k-z)
  \right],\\
    \mathcal{D}_m^{\prime}x_{m,\mathcal{D}_m^{\prime}}^{t+1}&=&-\mathcal{D}_m^{\prime}(\rho \mathcal{D}_m^{\prime\top} \mathcal{D}_m^{\prime})^{-1}
  \left[
  \lambda R_m^{\prime}(x_{m,\mathcal{D}_m^{\prime}}^{t+1})+\mathcal{D}_m^{\prime\top}y^t+\rho
  \mathcal{D}_m^{\prime\top}(\sum_{\substack
{k=1\\k\neq m}}^{M}\mathcal{D}_k\tilde{x}_k-z)
  \right].
\end{eqnarray*}

%\begin{eqnarray*}
%% \nonumber % Remove numbering (before each equation)
%  x_{m,\mathcal{D}_m}^{t+1}&=&-(\rho \mathcal{D}_m^\top \mathcal{D}_m)^{-1}
%  \left[
%  \lambda R_m^{\prime}(x_{m,\mathcal{D}_m}^{t+1})+y^t\mathcal{D}_m+\rho(\sum_{\substack
%{k=1\\k\neq m}}^{M}\mathcal{D}_kx_k-z)^\top \mathcal{D}_m
%  \right],\\
%    x_{m,\mathcal{D}_m^{\prime}}^{t+1}&=&-(\rho \mathcal{D}_m^{\prime\top} \mathcal{D}_m^{\prime})^{-1}
%  \left[
%  \lambda R_m^{\prime}(x_{m,\mathcal{D}_m^{\prime}}^{t+1})+y^t\mathcal{D}_m^{\prime}+\rho(\sum_{\substack
%{k=1\\k\neq m}}^{M}\mathcal{D}_kx_k-z)^\top \mathcal{D}_m^{\prime}
%  \right].
%\end{eqnarray*}

Therefore we have
\begin{eqnarray*}
&&\mathcal{D}_mx_{m,\mathcal{D}_m}^{t+1}-\mathcal{D}_m^{\prime}x_{m,\mathcal{D}_m^{\prime}}^{t+1}
\\
&&=-\mathcal{D}_m(\rho \mathcal{D}_m^\top \mathcal{D}_m)^{-1}
  \left[
  \lambda R_m^{\prime}(x_{m,\mathcal{D}_m}^{t+1})+\mathcal{D}_m^\top y^t\mathcal{D}_m+\rho\mathcal{D}_m^\top(\sum_{\substack
{k=1\\k\neq m}}^{M}\mathcal{D}_k\tilde{x}_k-z) 
  \right]\\
  &&~~~+\mathcal{D}_m^{\prime}(\rho \mathcal{D}_m^{\prime\top} \mathcal{D}_m^{\prime})^{-1}
  \left[
  \lambda R_m^{\prime}(x_{m,\mathcal{D}_m^{\prime}}^{t+1})+\mathcal{D}_m^{\prime\top}y^t+
  \rho\mathcal{D}_m^{\prime\top}(\sum_{\substack
{k=1\\k\neq m}}^{M}\mathcal{D}_k\tilde{x}_k-z)
  \right]\\
%% &&=[(\rho \mathcal{D}_m^\top \mathcal{D}_m)(\rho \mathcal{D}_m^{\prime\top} \mathcal{D}_m^{\prime})]^{-1}\\
%% &&~~~\times\left[(\rho \mathcal{D}_m^\top \mathcal{D}_m)\left(
%%\lambda R_m^{\prime}(x_{m,\mathcal{D}_m^{\prime}}^{t+1})+y^t\mathcal{D}_m^{\prime}+\rho(\sum_{\substack
%%{k=1\\k\neq m}}^{M}\mathcal{D}_kx_k-z)^\top \mathcal{D}_m^{\prime}
%% \right)\right.\\
%%&&~~~~~~-\left.
%%(\rho \mathcal{D}_m^{\prime\top} \mathcal{D}_m^{\prime})
%%\left(
%%\lambda R_m^{\prime}(x_{m,\mathcal{D}_m}^{t+1})+y^t\mathcal{D}_m+\rho(\sum_{\substack
%%{k=1\\k\neq m}}^{M}\mathcal{D}_kx_k-z)^\top \mathcal{D}_m
%%\right)
%%\right]\\
&&=\mathcal{D}_m(\rho \mathcal{D}_m^\top \mathcal{D}_m)^{-1}\\
&&~~~~~~\times\left[
\lambda (R_m^{\prime}(x_{m,\mathcal{D}_m^{\prime}}^{t+1})-R_m^{\prime}(x_{m,\mathcal{D}_m}^{t+1}))
+(\mathcal{D}_m^{\prime}-\mathcal{D}_m)^\top y^t+\rho(\mathcal{D}_m^{\prime}-\mathcal{D}_m)^\top(\sum_{\substack
{k=1\\k\neq m}}^{M}\mathcal{D}_k\tilde{x}_k-z)
\right]\\%%
&&~~~+[\mathcal{D}_m^{\prime}(\rho \mathcal{D}_m^{\prime\top} \mathcal{D}_m^{\prime})^{-1}
-
\mathcal{D}_m(\rho \mathcal{D}_m^{\top} \mathcal{D}_m)^{-1}
]\\
&&~~~~~~\times\left(
  \lambda R_m^{\prime}(x_{m,\mathcal{D}_m^{\prime}}^{t+1})+\mathcal{D}_m^{\prime\top} y^t+\rho\mathcal{D}_m^{\prime\top}(\sum_{\substack
{k=1\\k\neq m}}^{M}\mathcal{D}_k\tilde{x}_k-z)
\right).
%%% &&=[(\rho \mathcal{D}_m^\top \mathcal{D}_m)(\rho \mathcal{D}_m^{\prime\top} \mathcal{D}_m^{\prime})]^{-1}\\
%%%  &&~~~\times\left[\rho \mathcal{D}_m^\top \mathcal{D}_m\left(
%%%\lambda R_m^{\prime}(x_{m,\mathcal{D}_m^{\prime}}^{t+1})+y^t\mathcal{D}_m^{\prime}+\rho(\sum_{\substack
%%%{k=1\\k\neq m}}^{M}\mathcal{D}_kx_k-z)^\top \mathcal{D}_m^{\prime}
%%% \right)\right.\\
%%%&&~~~~~~-\left.
%%%\rho \mathcal{D}_m^{\prime\top} \mathcal{D}_m^{\prime}
%%%\left(
%%%\lambda R_m^{\prime}(x_{m,\mathcal{D}_m}^{t+1})+y^t\mathcal{D}_m+\rho(\sum_{\substack
%%%{k=1\\k\neq m}}^{M}\mathcal{D}_kx_k-z)^\top \mathcal{D}_m
%%%\right)
%%%\right]\\
%%%&&~~~+[(\rho \mathcal{D}_m^\top \mathcal{D}_m)(\rho \mathcal{D}_m^{\prime\top} \mathcal{D}_m^{\prime})]^{-1}\\
%%%&&~~~~~~\times \frac{1}{\eta_m^{t+1}}
%%%\left(
%%%y^t(\mathcal{D}_m^{\prime}-\mathcal{D}_m)+\rho(\sum_{\substack
%%%{k=1\\k\neq m}}^{M}\mathcal{D}_kx_k-z)^\top (\mathcal{D}_m^{\prime}-\mathcal{D}_m)
%%%\right)
\end{eqnarray*}
Denote
\begin{eqnarray*}
&&\Phi_1=\mathcal{D}_m(\rho \mathcal{D}_m^\top \mathcal{D}_m)^{-1}\\
&&~~~~~~\times\left[
\lambda (R_m^{\prime}(x_{m,\mathcal{D}_m^{\prime}}^{t+1})-R_m^{\prime}(x_{m,\mathcal{D}_m}^{t+1}))
+(\mathcal{D}_m^{\prime}-\mathcal{D}_m)^\top y^t+\rho(\mathcal{D}_m^{\prime}-\mathcal{D}_m)^\top(\sum_{\substack
{k=1\\k\neq m}}^{M}\mathcal{D}_k\tilde{x}_k-z)
\right],\\
%%%&&~~~~~~-
%%%\rho \mathcal{D}_m^{\prime\top} \mathcal{D}_m^{\prime}
%%%\left(
%%%\lambda\sum_{m=1}^{M}R_m^{\prime}(\tilde{x}_m^t)+y^t\mathcal{D}_m+\rho(\sum_{\substack
%%%{k=1\\k\neq m}}^{M}\mathcal{D}_kx_k-z)^\top \mathcal{D}_m-\frac{1}{\eta_m^{t+1}}\tilde{x}_m^t
%%%\right)\\
&&\Phi_2=[\mathcal{D}_m^{\prime}(\rho \mathcal{D}_m^{\prime\top} \mathcal{D}_m^{\prime})^{-1}
-
\mathcal{D}_m(\rho \mathcal{D}_m^{\top} \mathcal{D}_m)^{-1}
]\\
&&~~~~~~\times\left(
  \lambda R_m^{\prime}(x_{m,\mathcal{D}_m^{\prime}}^{t+1})+\mathcal{D}_m^{\prime\top}y^t+
  \rho\mathcal{D}_m^{\prime\top}(\sum_{\substack
{k=1\\k\neq m}}^{M}\mathcal{D}_k\tilde{x}_k-z) 
\right).
\end{eqnarray*}
As a result:
\begin{eqnarray}
\label{l2norm}
\mathcal{D}_mx_{m,\mathcal{D}_m}^{t+1}-\mathcal{D}_m^{\prime}x_{m,\mathcal{D}_m^{\prime}}^{t+1}=
\Phi_1+\Phi_2.
\end{eqnarray}
In the following, we will analyze the components in (\ref{l2norm}) term by term. The object
is to prove $\max_{\substack{\mathcal{D}_m,D_m^{\prime}\\
\|\mathcal{D}_m-D_m^{\prime}\|\leq1
}}
\|x_{m,\mathcal{D}_m}^{t+1}-x_{m,\mathcal{D}_m^{\prime}}^{t+1}\|$ is bounded.
To see this, notice that
\begin{eqnarray*}
&&\max_{\substack{\mathcal{D}_m,D_m^{\prime}\\
\|\mathcal{D}_m-D_m^{\prime}\|\leq1
}}
\|\mathcal{D}_mx_{m,\mathcal{D}_m}^{t+1}-\mathcal{D}_m^{\prime}x_{m,\mathcal{D}_m^{\prime}}^{t+1}\|\\
&&\leq\max_{\substack{\mathcal{D}_m,D_m^{\prime}\\
\|\mathcal{D}_m-D_m^{\prime}\|\leq1
}}\|\Phi_1\|
+\max_{\substack{\mathcal{D}_m,D_m^{\prime}\\
\|\mathcal{D}_m-D_m^{\prime}\|\leq1
}}\|\Phi_2\|.
\\
%%%&&=\max_{\substack{\mathcal{D}_m,D_m^{\prime}\\
%%%\|\mathcal{D}_m-D_m^{\prime}\|\leq1
%%%}}\|\Xi\|^{-1}\times
%%%\left(\max_{\substack{\mathcal{D}_m,D_m^{\prime}\\
%%%\|\mathcal{D}_m-D_m^{\prime}\|\leq1
%%%}}\|\Phi_1\|
%%%+\max_{\substack{\mathcal{D}_m,D_m^{\prime}\\
%%%\|\mathcal{D}_m-D_m^{\prime}\|\leq1
%%%}}\|\Phi_2\|
%%%\right)\\
\end{eqnarray*}
For $\max_{\substack{\mathcal{D}_m,D_m^{\prime}\\
\|\mathcal{D}_m-D_m^{\prime}\|\leq1
}}\|\Phi_2\|$,
from assumption \ref{theo:assumptions_pri_added}.\ref{item:assum_7_pri},
we have
\begin{eqnarray*}
  &&\max_{\substack{\mathcal{D}_m,D_m^{\prime}\\
\|\mathcal{D}_m-D_m^{\prime}\|\leq1
}}\|\Phi_2\|\\
  &&\leq
\left|\left|\frac{2}{d_m\rho}
\left(
  \lambda R_m^{\prime}(x_{m,\mathcal{D}_m^{\prime}}^{t+1})+\mathcal{D}_m^{\prime\top}y^t+
  \rho\mathcal{D}_m^{\prime\top}(\sum_{\substack
{k=1\\k\neq m}}^{M}\mathcal{D}_k\tilde{x}_k-z)
\right)\right|\right|.
\end{eqnarray*}
By mean value theorem, we have
\begin{eqnarray*}
&&\left|\left|\frac{2}{d_m\rho}
\left(
  \lambda\mathcal{D}_m^{\prime\top} R_m^{\prime\prime}(x_{\ast})+\mathcal{D}_m^{\prime\top}y^t+\rho\mathcal{D}_m^{\prime\top}(\sum_{\substack
{k=1\\k\neq m}}^{M}\mathcal{D}_k\tilde{x}_k-z)
\right)\right|\right|\\
&&\leq\frac{2}{d_m\rho}\left[\lambda\| R_m^{\prime\prime}(\cdot)\|
+\|y^t\|+\rho\|(\sum_{\substack
{k=1\\k\neq m}}^{M}\mathcal{D}_k\tilde{x}_k-z)\|
\right].
\end{eqnarray*}

For $\max_{\substack{\mathcal{D}_m,D_m^{\prime}\\
\|\mathcal{D}_m-D_m^{\prime}\|\leq1
}}\|\Phi_1\|$, we have

\begin{eqnarray*}
&&\max_{\substack{\mathcal{D}_m,D_m^{\prime}\\
\|\mathcal{D}_m-D_m^{\prime}\|\leq1
}}\|
\Phi_1
\|
\leq
\left|\left|
\mathcal{D}_m(\rho \mathcal{D}_m^\top \mathcal{D}_m)^{-1}\right.\right.\\
&&\left.\left.\times\left[
\lambda (R_m^{\prime}(x_{m,\mathcal{D}_m^{\prime}}^{t+1})-R_m^{\prime}(x_{m,\mathcal{D}_m}^{t+1}))
+(\mathcal{D}_m^{\prime}-\mathcal{D}_m)^\top y^t+\rho
(\mathcal{D}_m^{\prime}-\mathcal{D}_m)^\top (\sum_{\substack
{k=1\\k\neq m}}^{M}\mathcal{D}_k\tilde{x}_k-z)
\right]
\right|\right|\\
&&\leq
\rho^{-1}\|(\mathcal{D}_m^\top \mathcal{D}_m)^{-1}\|
\left[\lambda \|R_m^{\prime\prime}(\cdot)\|+\|y^t\|+\rho\|(\sum_{\substack
{k=1\\k\neq m}}^{M}\mathcal{D}_k\tilde{x}_k-z)^\top\|\right]\\
&&=\frac{1}{d_m\rho}\left[\lambda \|R_m^{\prime\prime}(\cdot)\|+\|y^t\|+\rho\|(\sum_{\substack
{k=1\\k\neq m}}^{M}\mathcal{D}_k\tilde{x}_k-z)\|\right].
\end{eqnarray*}

Thus by assumption \ref{theo:assumptions_pri_added}.\ref{item:assum_5_pri}-\ref{theo:assumptions_pri_added}.\ref{item:assum_6_pri}
\begin{eqnarray*}
&&\max_{\substack{\mathcal{D}_m,D_m^{\prime}\\
\|\mathcal{D}_m-D_m^{\prime}\|\leq1
}}
\|\mathcal{D}_mx_{m,\mathcal{D}_m}^{t+1}-\mathcal{D}_m^{\prime}x_{m,\mathcal{D}_m^{\prime}}^{t+1}\|\\
&&\leq\frac{3}{d_m\rho}\left[\lambda c_1+\|y^t\|+\rho\|(\sum_{\substack
{k=1\\k\neq m}}^{M}\mathcal{D}_k\tilde{x}_k-z)^\top\|\right]\\
&&\leq\frac{3}{d_m\rho}\left[\lambda c_1+\|y^t\|+\rho\|z\|
+\rho\sum_{\substack
{k=1\\k\neq m}}^{M}\|\tilde{x}_k\|\right]\\
&&\leq\frac{3}{d_m\rho}\left[\lambda c_1+(1+M\rho)b_1\right]
\end{eqnarray*}
is bounded.
\hfill$\square$

\subsection{Proof of Theorem~\ref{theo:DP}}

{\it Proof:} The privacy loss from $D_m\tilde{x}_m^{t+1}$ is calculated by:

\begin{eqnarray*}
  \left|
  \text{ln}\frac{P(\mathcal{D}_m\tilde{x}_m^{t+1}|\mathcal{D}_m)}{P(\mathcal{D}_m^{\prime}\tilde{x}_m^{t+1}|\mathcal{D}_m^{\prime})}
  \right|=
  \left|
  \text{ln}\frac{P(\mathcal{D}_m\tilde{x}_{m,\mathcal{D}_m}^{t+1}+\mathcal{D}_m\xi_m^{t+1})}
  {P(\mathcal{D}_m^{\prime}\tilde{x}_{m,\mathcal{D}_m^{\prime}}^{t+1}+\mathcal{D}_m^{\prime}\xi_m^{\prime,t+1})}
  \right|=
    \left|
  \text{ln}\frac{P(\mathcal{D}_m\xi_m^{t+1})}
  {P(\mathcal{D}_m^{\prime}\xi_m^{\prime,t+1})}
  \right|.
\end{eqnarray*}

Since $\mathcal{D}_m\xi_m^{t+1}$ and $\mathcal{D}_m^{\prime}\xi_m^{\prime,t+1}$ are sampled from $\mathcal{N}(0,\sigma_{m,t+1}^2)$,
combine with lemma \ref{lemma:privacy}, we have
\begin{eqnarray*}
&&\left|
  \text{ln}\frac{P(\mathcal{D}_m\xi_m^{t+1})}
  {P(\mathcal{D}_m^{\prime}\xi_m^{\prime,t+1})}
  \right|\\
&&=\left|
\frac{2\xi_m^{t+1}\|\mathcal{D}_mx_{m,\mathcal{D}_m}^{t+1}-\mathcal{D}_m^{\prime}x_{m,\mathcal{D}_m^{\prime}}^{t+1}\|+
\|\mathcal{D}_mx_{m,\mathcal{D}_m}^{t+1}-\mathcal{D}_m^{\prime}x_{m,\mathcal{D}_m^{\prime}}^{t+1}\|^2}
{2\sigma_{m,t+1}^2}
\right|\\
&&\leq
\left|
\frac{2\mathcal{D}_m\xi_m^{t+1}\mathbb{C}+\mathbb{C}^2}{2\frac{\mathbb{C}^2\cdot2\text{ln}(1.25/\sigma)}{\varepsilon^2}}
\right|\\
&&=\left|
\frac{(2\mathcal{D}_m\xi_m^{t+1}+\mathbb{C})\varepsilon^2}{4\mathbb{C}\text{ln}(1.25/\sigma)}
\right|.
\end{eqnarray*}
In order to make $\left|
\frac{(2\mathcal{D}_m\xi_m^{t+1}+\mathbb{C})\varepsilon^2}{4\mathbb{C}\text{ln}(1.25/\sigma)}
\right|\leq \varepsilon$, we need to make sure
\begin{eqnarray*}
  \left|
  \mathcal{D}_m\xi_m^{t+1}
  \right|
  \leq
  \frac{2\mathbb{C}\text{ln}(1.25/\sigma)}{\varepsilon}-\frac{\mathbb{C}}{2}.
\end{eqnarray*}

In the following, we need to proof
\begin{eqnarray}
\label{prbinq}
  P(\left|
  \mathcal{D}_m\xi_m^{t+1}
  \right|
  \geq
  \frac{2\mathbb{C}\text{ln}(1.25/\sigma)}{\varepsilon}-\frac{\mathbb{C}}{2})
  \leq\delta
\end{eqnarray}
holds. However, we will proof a stronger result that lead to
(\ref{prbinq}). Which is
\begin{eqnarray*}
   P(
  \mathcal{D}_m\xi_m^{t+1}
  \geq
  \frac{2\mathbb{C}\text{ln}(1.25/\sigma)}{\varepsilon}-\frac{\mathbb{C}}{2})
  \leq\frac{\delta}{2}.
\end{eqnarray*}

Since the tail bound of normal distribution $\mathcal{N}(0,\sigma_{m,t+1}^2)$ is:
\begin{eqnarray*}
  P(\mathcal{D}_m\xi_m^{t+1}>r)\leq\frac{\sigma_{m,t+1}}{r\sqrt{2\pi}}e^{-\frac{r^2}{2\sigma_{m,t+1}^2}}.
\end{eqnarray*}
Let $r=\frac{2\mathbb{C}\text{ln}(1.25/\sigma)}{\varepsilon}-\frac{\mathbb{C}}{2}$, we then have
\begin{eqnarray*}
  &&P(
  \mathcal{D}_m\xi_m^{t+1}
  \geq
  \frac{2\mathbb{C}\text{ln}(1.25/\sigma)}{\varepsilon}-\frac{\mathbb{C}}{2})\\
  &&\leq\frac{\mathbb{C}\sqrt{2\text{ln}(1.25/\sigma)}}{r\sqrt{2\pi}\varepsilon}
  \exp\left[
  -\frac{[4\text{ln}(1.25/\sigma)-\varepsilon]^2}{8\text{ln}(1.25/\sigma)}
  \right].
\end{eqnarray*}
When $\delta$ is small and let $\varepsilon\leq 1$, we then have
\begin{eqnarray}
\label{interme1}
  \frac{\sqrt{2\text{ln}(1.25/\sigma)}2}{(4\text{ln}(1.25/\sigma)-\varepsilon)\sqrt{2\pi}}
  \leq\frac{\sqrt{2\text{ln}(1.25/\sigma)}2}{(4\text{ln}(1.25/\sigma)-1)\sqrt{2\pi}}
  <\frac{1}{\sqrt{2\pi}}.
\end{eqnarray}
As a result, we can proof that
\begin{eqnarray*}
-\frac{[4\text{ln}(1.25/\sigma)-\varepsilon]^2}{8\text{ln}(1.25/\sigma)}<
\text{ln}(\sqrt{2\pi}\frac{\delta}{2}
)
\end{eqnarray*}
by equation (\ref{interme1}). Thus we have
\begin{eqnarray*}
P(
  \mathcal{D}_m\xi_m^{t+1}
  \geq
  \frac{2\mathbb{C}\text{ln}(1.25/\sigma)}{\varepsilon}-\frac{\mathbb{C}}{2})<
  \frac{1}{\sqrt{2\pi}}\exp(\text{ln}(\sqrt{2\pi}\frac{\delta}{2})=\frac{\delta}{2}.
\end{eqnarray*}

Thus we proved (\ref{prbinq}) holds. Define
\begin{eqnarray*}
&&\mathbb{A}_1=\{
\mathcal{D}_m\xi_m^{t+1}:|\mathcal{D}_m\xi_m^{t+1}|\leq\frac{1}{\sqrt{2\pi}}\exp(\text{ln}(\sqrt{2\pi}\frac{\delta}{2}
\},\\
&&\mathbb{A}_2=\{
\mathcal{D}_m\xi_m^{t+1}:|\mathcal{D}_m\xi_m^{t+1}|>\frac{1}{\sqrt{2\pi}}\exp(\text{ln}(\sqrt{2\pi}\frac{\delta}{2}
\}.
\end{eqnarray*}
Thus we obtain the desired result:
\begin{eqnarray*}
&&P(\mathcal{D}_m^{\prime}\tilde{x}_m^{t+1}|\mathcal{D}_m)\\
&&=
P(\mathcal{D}_mx_{m,\mathcal{D}_m}^{t+1}+\mathcal{D}_m\xi_m^{t+1}:\mathcal{D}_m\xi_m^{t+1}\in\mathbb{A}_1)\\
&&+P(\mathcal{D}_mx_{m,\mathcal{D}_m}^{t+1}+\mathcal{D}_m\xi_m^{t+1}:\mathcal{D}_m\xi_m^{t+1}\in\mathbb{A}_2)\\
&&<e^{\varepsilon}P(\mathcal{D}_mx_{m,\mathcal{D}_m^{\prime}}^{t+1}
+\mathcal{D}_m\xi_m^{\prime,t+1})+\delta=e^{\varepsilon}P(\mathcal{D}_m\tilde{x}_m^{t+1}|\mathcal{D}_m^{\prime})+\delta.
\end{eqnarray*}
\hfill$\square$ 
\subsection{Implementation}
The implementation of the prototype system can be found at \url{https://www.dropbox.com/sh/imfz3k2fhf3zkrc/AACcbREvj_PBSjEUdZmBfoFUa?dl=0}.

\end{document}